\documentclass[letterpaper]{article}
\usepackage[preprint]{aaai2027}
\usepackage[hyphens]{url}
\usepackage{graphicx}
\usepackage{natbib}
\usepackage{caption}
\usepackage{booktabs}
\usepackage{float}
\usepackage{amsmath}
\usepackage{amssymb}
\usepackage[most]{tcolorbox}
\usepackage{microtype}
\newtcolorbox{promptbox}{
  colback=gray!7,
  colframe=black!25,
  boxrule=0.35pt,
  arc=1pt,
  left=4pt,
  right=4pt,
  top=3pt,
  bottom=3pt,
  fontupper=\small,
  breakable
}

\title{Evidence Interfaces Shape How Retrieval-Augmented Readers Use Support}
\author{Junchi Liao, Jiawen Deng, Fuji Ren$^{*}$}
\affiliations{$^{*}$Corresponding author}

\begin{document}
\maketitle

\begin{abstract}
In multi-hop RAG evaluation, a top-$k$ answer score can hide two
different failures: the retrieval window may drop part of the support chain, or
it may contain support in a form the adapted reader does not use well. We call
this reader-facing form of retrieved evidence an \emph{evidence interface}.
Using three support-annotated multi-hop QA benchmarks, we compare matched
adapted readers trained with raw context, retrieval windows, and gold-support
diagnostic renderings. These comparisons distinguish support-availability
failures from remaining reader--interface effects. Top-$k$ windows become interpretable only after checking
whether the complete annotated support chain survives: when it does, short
ranked windows can match or improve over raw context; when it does not, missing
support explains much of the loss. Gold support-first improves matched readers;
on 2Wiki and MuSiQue, a support-supervised ranker raises coverage and recovers
raw-context quality at lower prompt cost, while retaining gold headroom.
Support-removal checks further show that the gains rely on exposed evidence,
not only answer priors. On support-annotated evaluations, top-$k$ answer scores
should therefore be reported together with complete-support coverage.
\end{abstract}

\section{Introduction}

Consider a HotpotQA example that asks for the lead character in the 1960s
sitcom Get Smart. The raw evidence pool contains the needed support
titles, Get Smart, Again! and Barbara Feldon, but they are
separated by distractors. A reader adapted on the raw context answers the
nearby entity Agent 99. When the same support is moved to the
front of the input, with distractors still present, a matched reader answers
Maxwell Smart. The evidence was available in both inputs. The candidate pool
did not change; the reader-facing presentation did. This is not a simple
retrieval miss. It is a case where available evidence is harder or easier for
the reader to use. Appendix D shows the rendered case.

This is a reader-side measurement problem for retrieval-augmented generation
(RAG) \cite{lewis2020rag}. Standard RAG evaluation naturally asks whether the
retriever found relevant evidence and whether the reader produced the right
answer. That view is necessary: if a multi-hop support chain is absent from the
reader input, the answer may be impossible. But it is not sufficient. The
reader consumes a concrete input string, not an abstract relevance set. The
same support can be buried in raw context, reordered by a ranker, truncated
into a top-$k$ window, moved before distractors, or converted into a compact
structure. These choices can change what an adapted reader learns from the
same question-answer pairs. Ignoring this reader-facing form can make
aggregate answer scores hard to interpret.

The ambiguity is sharpest for top-$k$ RAG scores. A short retrieval window can
lower prompt cost and remove distractors, but it can also drop one support unit
needed for the answer. A lower F1 score can mean that the ranker failed to
preserve the support chain, or that the reader failed to use a surviving chain.
A higher F1 score can come from better use of the presented evidence rather than better
retrieval. The practical consequence is that a system designer may optimize the
wrong component: changing the retriever when the reader needs a more usable
evidence form, or changing the reader when the retrieval window is missing
support. Thus the attribution of aggregate top-$k$ F1 is ambiguous unless it is paired with
annotated-support survival. Long-context work already shows that evidence
position affects model behavior \cite{liu2023lost}; here we ask how the
evidence presentation handed to an adapted RAG reader changes the learned
reader.

\begin{figure*}[t]
\centering
\includegraphics[width=\textwidth]{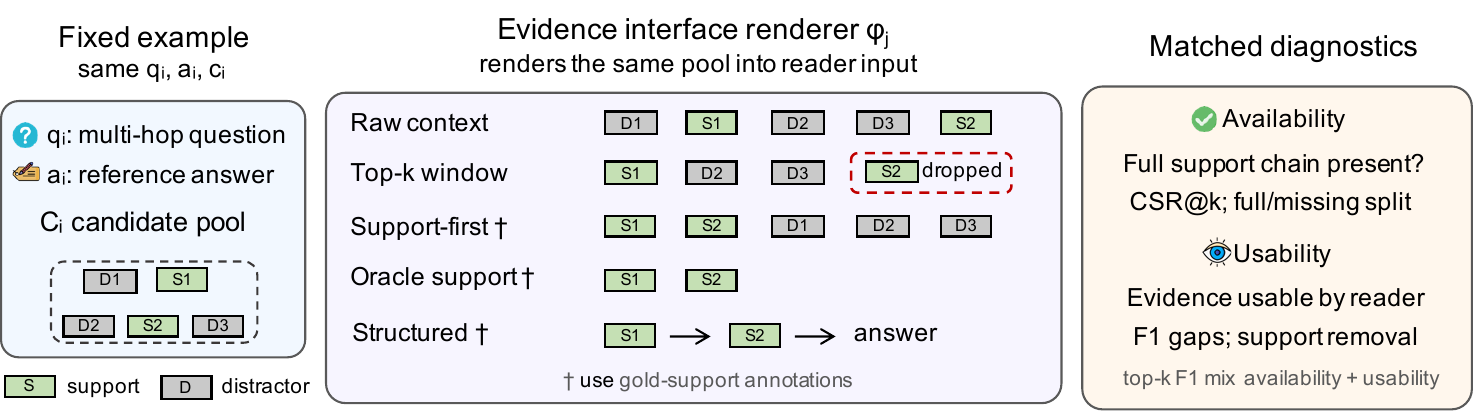}
\caption{Overview of evidence-interface diagnostics. For a fixed question,
answer, and candidate pool, an interface renderer turns the same evidence into
different reader inputs. The matched diagnostics then separate support
availability from residual reader--interface effects; daggered interfaces use gold-support
annotations.}
\label{fig:aaai-overview}
\end{figure*}

We call this reader-facing presentation an evidence interface. An
interface maps a question, a candidate evidence pool, optional ranking output,
and optional support annotations to the actual reader input. This framing lets
us separate two failure sources that answer accuracy often mixes: support
availability, whether the selected input contains every annotated support unit, and
remaining reader--interface effects, whether answer quality changes even when
the annotated support survives in the rendered input. We
measure both before interpreting whether a top-$k$ retrieval window helps or
hurts the reader.
Figure~\ref{fig:aaai-overview} summarizes this diagnostic view.

We study this distinction on three support-annotated multi-hop QA benchmarks
\cite{yang2018hotpotqa,ho2020twowiki,trivedi2022musique}. For each interface,
we train matched adapted readers with the same questions, answers, reader
family, training-example budget, adapter capacity, and validation examples. The interfaces include raw
context, retrieval windows, and gold-support diagnostic renderings. The
gold-support rows are not serving methods; they give reference conditions for
asking what remains possible when the needed support is present and organized
for the reader.

The results make the top-$k$ ambiguity concrete. When the ranker-specific @5 window
keeps the complete annotated support chain, it is tied with raw context or
better; when support is missing, F1 drops sharply. Gold support-first renderings
improve the matched readers, and fixed-adapter restore/delete checks show that
the effect depends on the exposed evidence. Support-supervised rankers recover
substantially more of this practical gap than a weak lexical selector, while
remaining below the gold diagnostic. Scaling, rank, and cross-reader checks
keep the claim bounded: this is a diagnostic study of support-annotated
multi-hop QA with closed candidate pools, not a new state-of-the-art QA system.

This paper makes three contributions:
\begin{itemize}
\item We formalize evidence interfaces as a reader-facing variable and compare
them with matched per-interface adaptation rather than inference-only prompt
swaps.
\item We combine complete-support coverage with fixed-adapter restore/delete
interventions to separate missing evidence from evidence that the reader uses
poorly in its presented form.
\item We show that aggregate @5 scores can average over opposite
coverage-conditioned effects, establishing complete-support coverage as a
necessary companion metric on support-annotated multi-hop benchmarks.
\end{itemize}

\section{Evidence Interface Diagnostics}

We use evidence interface diagnostics as a measurement framework for adapted
RAG readers. The framework treats the reader-facing evidence form as a
controlled intervention, then asks whether failures come from missing support
or from support that is present but difficult for the reader to use. It does
not require a new retriever or reader architecture.

\subsection{Evidence Interfaces}

Consider a multi-hop question whose answer requires two support units,
$S_1$ and $S_2$. A raw-context input may contain both units but place them
after distractors and far apart. A top-$k$ window may be shorter but keep only
$S_1$. A support-first diagnostic may keep the same candidate pool while
moving $S_1$ and $S_2$ to the front. An oracle-support diagnostic may show
only the two support units. These are different reader inputs for the same
question-answer example.

Here, an interface denotes the reader-facing evidence form, not the retriever
itself. Formally, each interface is a renderer: it selects,
orders, and optionally restructures evidence before the shared prompt template
is applied. For example
$i$, let $q_i$ be the question, $a_i$ the reference answer, and
$C_i=(c_{i1},\ldots,c_{im})$ the candidate evidence pool. When support
annotations exist, let $S_i\subseteq C_i$ denote the candidate units identified
as support. Sentence-level facts are lifted to their title-associated
paragraphs for selection and CSR; oracle renderings may use the finer-grained
annotations. The experimental setup gives the dataset-specific mapping. A ranker may also produce an ordering
$R_i=R(q_i,C_i)$. We write interface $j$ as
\begin{equation}
E_i^{(j)}=\phi_j(q_i,C_i,R_i,S_i),\qquad
x_i^{(j)}=T(q_i,E_i^{(j)}),
\label{eq:aaai-interface}
\end{equation}
where $E_i^{(j)}$ is the rendered evidence shown by interface $j$ and
$T(\cdot)$ is the shared prompt template.

The comparison fixes the question, answer, and candidate pool, and changes the
rendered evidence. Token count is not held fixed, because length is part of the
interface. The full adaptation protocol matches reader family, training-example
count, epochs, adapter capacity, and evaluation examples; total processed tokens
can differ because interface lengths differ.

Table~\ref{tab:aaai-interface-families} summarizes the six interface families.
Retrieval windows use BM25 or cross-encoder rankers
\cite{robertson2009bm25,nogueira2019bert}. Gold-support rows use annotations
or gold structure and are diagnostics rather than serving inputs.

\begin{table}[h]
\centering
\caption{Evidence-interface families and their experimental roles.}
\label{tab:aaai-interface-families}
\small
\setlength{\tabcolsep}{2.5pt}
\begin{tabular}{@{}p{0.25\columnwidth}p{0.20\columnwidth}p{0.18\columnwidth}p{0.27\columnwidth}@{}}
\toprule
Interface & Gold & Distractors & Role \\
\midrule
No context & -- & -- & answer prior \\
Raw context & No & Yes & long context \\
Top-$k$ window & No & Variable & budget/coverage \\
Support-first & Yes & Yes & placement probe \\
Oracle support & Yes & No & clean support \\
Gold-chain triples & Yes & No & answer-exposing ceiling \\
\bottomrule
\end{tabular}
\end{table}

The families cover three interventions that a single answer score can mix:
selecting support, changing its position among distractors, and compacting its
representation. No-context and raw-context rows anchor the comparison, while
retrieval windows test budgeted evidence selection. Gold-support rows are
diagnostic reference conditions. The structured row is used only for
2WikiMultiHopQA and is treated as a ceiling because its triples often expose
the answer-bearing chain.

\subsection{Availability and Residual Reader--Interface Effects}

A RAG example can fail in two ways. The selected input may not contain every
annotated support unit; this is an availability failure. Or a residual
reader--interface effect may remain when support is present, because it is
placed after distractors, split across long text, or represented at an
inconvenient granularity. We measure this residual without treating it as a
separately identified latent mechanism. A single
top-$k$ answer score can mix these cases.

The two failure modes require different diagnostics. Availability can be
measured directly when support annotations exist: every annotated support unit
must be selected into the rendered evidence window. This selected-unit check is
an annotation-level proxy; Appendix C audits final tokenized prompt visibility
separately. The residual requires a matched comparison after support availability
is accounted for. If a top-$k$ window loses accuracy only
on examples where it misses support, the main problem is retrieval coverage.
If it still differs from raw context on examples where annotated support is
present, the reader-side interface remains part of the explanation.

\subsection{Diagnostic Protocol}

The protocol has three components: matched adaptation, coverage-conditioned
analysis, and evidence-removal checks. Raw context and retrieval windows are
deployable-style inputs. Support-first, oracle support, and gold-chain triples use
gold annotations or gold structure; they are counterfactual diagnostics, not
serving proposals. Their role is to estimate whether failures remain when the
needed support is present and easier for the reader to use.

For each interface $j$, we train a separate reader adapter on prompts
$x_i^{(j)}$ and answer targets $a_i$. The objective is the supervised answer
loss
\begin{equation}
\mathcal{L}_j(\Delta\theta)=
-\frac{1}{N}\sum_{i=1}^{N}
\log p_{\theta_0+\Delta\theta}(a_i\mid x_i^{(j)}),
\label{eq:aaai-loss}
\end{equation}
where $\theta_0$ is the frozen base reader and $\Delta\theta$ is the adapter
update. The main experiments use rank-8 LoRA adapters \cite{hu2022lora}. LoRA
is a measurement tool here: it fixes the trainable parameter count and number
of training examples so that interface contrasts are not confounded with
different adapter sizes or supervision counts.

All matched comparisons use the same reader family, train size, validation
questions, decoding rule, and answer target. The reader is retrained per
interface because each interface changes the training distribution. Thus, an
interface score measures matched train/evaluation performance for that
evidence form: it combines how easily the reader learns from the form and how
well it uses the same form at evaluation time. It is not a pure inference-time
prompt-swap effect; Appendix C reports cross-interface swaps that make this
boundary explicit.
The main interface contrast is the matched F1 difference against raw context on
the same validation questions:
\begin{equation}
\Delta_{j-\mathrm{raw}}=
\mathrm{F1}(I_j)-\mathrm{F1}(I_{\mathrm{raw}}).
\end{equation}

For retrieval windows, we measure complete-support recall:
\begin{equation}
\mathrm{CSR@}k=\frac{1}{N}\sum_i
\mathbf{1}\!\left[S_i\subseteq E_i^{\mathrm{top}k}\right].
\label{eq:aaai-csr}
\end{equation}
CSR@$k$ is an annotation-level selected-support check. It does not guarantee
that every support token remains visible after prompt rendering and truncation;
final-input visibility is audited separately. Unlike mean support fraction,
which gives partial credit for selecting only part of a chain, CSR@$k$ equals
one only when every annotated support unit is selected. We use CSR@$k$ to separate
coverage failures from reader-interface failures. We also use no-support
diagnostics: annotated support is removed from an
evidence-bearing interface and the reader is re-evaluated. The comparison tests
whether interface gains depend on exposed evidence rather than answer priors or
format alone.

\section{Experimental Setup}
\label{sec:aaai-exp-setup}

\subsection{Datasets and Evidence Units}

We evaluate on HotpotQA, 2WikiMultiHopQA, and MuSiQue
\cite{yang2018hotpotqa,ho2020twowiki,trivedi2022musique}. These datasets are
useful for this study because multi-hop answers are annotated with a support chain rather
than one isolated passage. HotpotQA provides supporting documents or sentences
within a distractor setting. 2WikiMultiHopQA provides annotated reasoning
paths, which also support a structured-evidence diagnostic. MuSiQue composes
multiple single-hop questions and is useful for testing whether short
retrieval windows preserve the annotated chain.

For raw, ranked, top-$k$, and support-first inputs, HotpotQA and
2WikiMultiHopQA use complete title-associated paragraphs as candidate units;
sentence-level supporting facts identify which paragraph titles are support,
but support-first moves the whole corresponding paragraphs. MuSiQue uses and
moves complete annotated paragraphs. Oracle support is intentionally more
compact: it keeps annotated supporting sentences for HotpotQA and 2Wiki, and
annotated paragraphs for MuSiQue. CSR checks support titles in the first two
datasets and supporting paragraphs in MuSiQue. Support recall is therefore a within-dataset diagnostic,
not a claim that units have identical granularity across benchmarks. In our
retrieval setting, the datasets span different annotated-support coverage
regimes under the rankers evaluated in the main matrix: HotpotQA is relatively
coverage-rich under the fine-tuned cross-encoder @5 window, 2WikiMultiHopQA
provides both support annotations and structured reasoning paths, and MuSiQue
is more fragile under the cross-encoder @5 window. These differences should
not be read as an intrinsic ranker-quality ordering across datasets.
Raw-context truncation is rare for HotpotQA and 2Wiki in our
validation audit, but occurs in 140 of 1000 MuSiQue prompts. Appendix C audits
support visibility after final tokenization; after filtering to examples where
support text is visible in both raw and support-first prompts, the
support-first gain remains positive on all datasets.

\subsection{Readers, Retrieval, and Metrics}

We use Qwen2.5-Coder-14B-Instruct with 4000 training examples as the main
matched setting because it is the common 14B reader for which all three
datasets and core interface families were completed under the same protocol.
The main adapters use rank-8 LoRA,
and decoding is fixed across interfaces within each reader setting. Scaling,
rank, and cross-reader checks are reported separately so that the main
interface comparison remains paired. The same interface families in
Table~\ref{tab:aaai-interface-families} are applied across datasets, with only
dataset-specific support units changing.

We report token-level answer F1 in the main paper. For retrieval windows, we
also report CSR and average prompt-token cost. Main @5 rows always name the
ranker in the table: FT-CE for HotpotQA and CE for 2Wiki and MuSiQue. The
support-removal sweep uses BM25@5 so the same non-oracle retrieval row is
available across datasets and readers. Accordingly, cross-dataset @5 numbers
describe the evaluated ranker regimes, not an intrinsic dataset-difficulty or
ranker-quality ordering.
Paired bootstrap intervals are used for matched result summaries in
the full experiments. Detailed rendering, retrieval, and adaptation settings are
in Appendix B, and additional results are in Appendix C.

\section{Main Results and Analysis}

We organize the analysis in four steps. First, we test whether changing the
evidence interface changes adapted-reader accuracy under matched training
conditions. Second, we split retrieval-window results by annotated-support
coverage. Third, we separate evidence ordering from truncation with secondary
controls. Finally, we check whether the effects survive data scaling, reader
changes, adapter-rank changes, and support removal.

\subsection{Interface Choice Changes Adapted Readers}

We first fix the question set, answers, reader family, training-example count,
epochs, and trainable parameter count, then vary only the evidence form.
Table~\ref{tab:aaai-main-matrix} gives the matched Qwen2.5-Coder-14B
comparison at train size 4000.

\paragraph{Changing the evidence form changes adapted readers.}
Support-first improves over raw context on all three datasets, with gains of
0.026 F1 on HotpotQA, 0.039 on 2Wiki, and 0.070 on MuSiQue. Oracle support is
highest on 2Wiki and MuSiQue, while the HotpotQA oracle row is close to
support-first. Oracle and gold-chain-triple rows are diagnostics: they mark remaining
headroom when support is isolated or already resolved into a compact chain.

The paired uncertainty estimates support this reading. Support-first is
positive over raw in all three datasets, with paired bootstrap intervals
excluding zero on 2Wiki and MuSiQue and narrowly positive on HotpotQA. Larger
oracle-support gains on 2Wiki and MuSiQue indicate that removing distractors can
matter beyond simply moving support earlier; on HotpotQA, the oracle interval
overlaps raw, so we do not treat it as a separate significant gain.

\begin{table}[h]
\centering
\scriptsize
\setlength{\tabcolsep}{2.5pt}
\caption{Main Qwen2.5-Coder-14B interface matrix (eval $n=300$). The @5 column uses the
ranker named in the second column: FT-CE for HotpotQA and CE for 2Wiki and MuSiQue. Daggered columns use
gold annotations and are diagnostics, not deployable rows; Gold tri. is a 2Wiki
gold-chain ceiling.}
\label{tab:aaai-main-matrix}
\begin{tabular}{llccccc}
\toprule
Dataset & Ranker & Raw & @5 & SF$^\dagger$ & Oracle$^\dagger$ & Gold tri.$^\dagger$ \\
\midrule
HotpotQA & FT-CE & 0.780 & 0.768 & 0.806 & 0.796 & -- \\
2Wiki & CE & 0.782 & 0.663 & 0.822 & 0.844 & 0.993 \\
MuSiQue & CE & 0.581 & 0.459 & 0.651 & 0.747 & -- \\
\bottomrule
\end{tabular}
\end{table}

\begin{table}[h]
\centering
\tiny
\setlength{\tabcolsep}{1.6pt}
\caption{Paired bootstrap intervals for Qwen2.5-Coder-14B at train size 4000
and eval $n=300$.
Values are F1 differences against raw context; @5 uses the ranker named in
Table~\ref{tab:aaai-main-matrix}. Intervals use 10,000 paired resamples over
matched validation ids.}
\label{tab:aaai-paired-uncertainty}
\begin{tabular}{lccc}
\toprule
Dataset & SF$-$raw & @5$-$raw & Oracle$-$raw \\
\midrule
HotpotQA & +.026 [+.001,+.053] & -.012 [-.045,+.021] & +.016 [-.020,+.053] \\
2Wiki & +.039 [+.007,+.073] & -.120 [-.170,-.070] & +.061 [+.020,+.104] \\
MuSiQue & +.070 [+.034,+.106] & -.121 [-.177,-.066] & +.166 [+.122,+.211] \\
\bottomrule
\end{tabular}
\end{table}

\begin{figure*}[t]
\centering
\includegraphics[width=\textwidth]{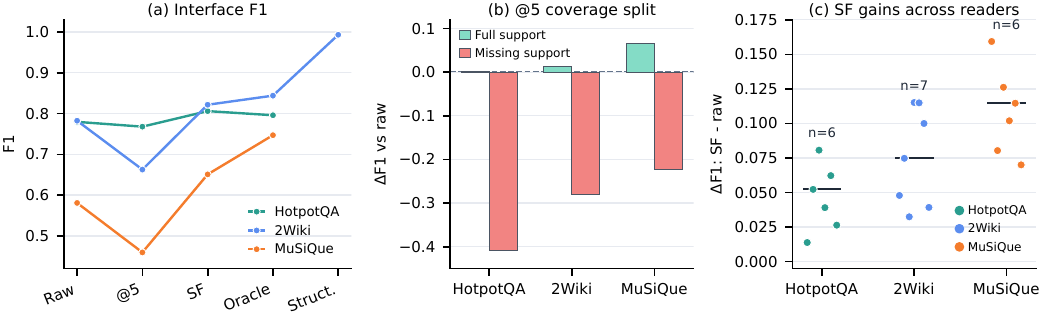}
\caption{Main evidence-interface effects. (a) Matched Qwen2.5-Coder-14B F1 across
interface families; gold-chain triples are available only for 2Wiki and form an
answer-exposing ceiling. (b) On the expanded 1000-example answer evaluation, the ranker-specific @5
losses are small when annotated support is selected and large when support is
missing. (c) Support-first gains are positive across available reader
conditions.}
\label{fig:aaai-interface-summary}
\end{figure*}

\paragraph{Retrieval windows are not uniformly better compression.}
The @5 row is close to raw on HotpotQA but loses roughly 0.10--0.13 F1 on
2Wiki and MuSiQue across the 300- and 1000-example views, despite reducing prompt tokens. This sign change is
why we treat retrieval windows as both coverage and interface interventions; the
next subsection separates the two.

\subsection{Retrieval Windows: Coverage vs. Usability}

Top-$k$ evaluation conflates two failure modes. A retrieval window can fail
because the ranker omits an annotated support unit, or because the reader fails
to use support that is present. Table~\ref{tab:aaai-retrieval-cost} summarizes
the tradeoff. The ranker-specific @5 window cuts prompt tokens by 41--72\%, but
annotated-support selection varies sharply across the evaluated ranker regimes.
The same five-document interface can therefore act like useful compression in
one setting and a coverage bottleneck in another.

\begin{table}[h]
\centering
\scriptsize
\setlength{\tabcolsep}{3pt}
\caption{Ranker-specific @5 retrieval audit and answer gap on 1000 examples at train size 4000. CSR@5, mean support, token cut, and @5--raw use the same validation slice.}
\label{tab:aaai-retrieval-cost}
\begin{tabular}{llcccc}
\toprule
Dataset & Ranker & CSR@5 (\%) & Mean sup. (\%) & Tok. cut (\%) & @5--raw \\
\midrule
HotpotQA & FT-CE & 94.8 & 97.4 & 53.6 & -0.019 \\
2Wiki & CE & 59.7 & 82.4 & 41.1 & -0.104 \\
MuSiQue & CE & 31.3 & 65.1 & 71.3 & -0.132 \\
\bottomrule
\end{tabular}
\end{table}

\begin{figure}[t]
\centering
\includegraphics[width=\columnwidth]{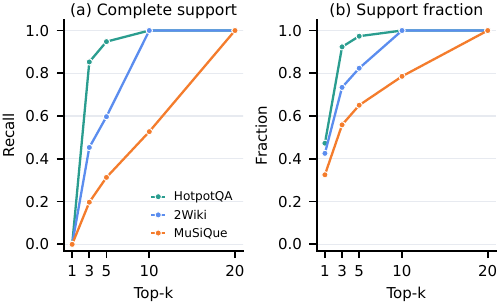}
\caption{Support recall of the main retrieval window used for each dataset.
CSR asks whether the full annotated chain is selected; support
fraction gives partial-chain coverage.}
\label{fig:aaai-retrieval-recall}
\end{figure}

Figure~\ref{fig:aaai-retrieval-recall} shows the corresponding support-recall
curves for the main ranker used in each dataset. Complete-support recall is the
stricter quantity: many windows contain part of the chain, but a multi-hop
reader can still miss the unit that completes the answer.

\begin{table}[h]
\centering
\scriptsize
\setlength{\tabcolsep}{3pt}
\caption{Qwen2.5-Coder-14B BM25 window-size check at train size 2000. Values are F1; SF uses gold support and is diagnostic, not deployable. BM25@10 is effectively the full ranked pool for HotpotQA and 2Wiki.}
\label{tab:aaai-bm25-window-size}
\begin{tabular}{lccccc}
\toprule
Dataset & Raw & BM25@5 & BM25@10 & SF$^\dagger$ & Tok.@10 \\
\midrule
HotpotQA & 0.748 & 0.690 & 0.773 & 0.801 & 1479 \\
2Wiki & 0.694 & 0.629 & 0.696 & 0.772 & 1046 \\
MuSiQue & 0.562 & 0.377 & 0.472 & 0.617 & 1318 \\
\bottomrule
\end{tabular}
\end{table}

\paragraph{A larger window helps, but does not erase the interface gap.}
Table~\ref{tab:aaai-bm25-window-size} compares BM25@5 with BM25@10 using the
same reader and train size. Increasing the window recovers much of the BM25@5
loss on HotpotQA and 2Wiki, where BM25@10 is close to the full ranked pool.
MuSiQue remains far below raw even after doubling the window. Larger non-oracle
windows reduce the coverage bottleneck, but they do not make support
organization irrelevant.

The curve shape matters. HotpotQA reaches high complete-support recall by
@5, so its FT-CE window mostly tests whether a shorter context is usable.
2Wiki has high partial support but many incomplete chains at @5, while MuSiQue
needs a much larger window before complete chains become common. Mean support
fraction is therefore not enough for multi-hop interpretation: the missing unit
is often the final hop that makes the answer identifiable.

The gap between partial and complete support is itself diagnostic. At @5,
mean support fraction versus complete-chain recall is 0.974 versus 0.948 on
HotpotQA, 0.824 versus 0.597 on 2Wiki, and 0.651 versus 0.313 on MuSiQue.
Thus MuSiQue often retrieves several relevant documents while still omitting
one decisive unit. Even at @10, its fraction reaches 0.786 but complete
support only 0.527. Extra context therefore repairs partial chains before it
reliably restores the full chain, explaining why token savings and
answerability can diverge under the same top-$k$ interface.

Table~\ref{tab:aaai-coverage-decomp} together with
Figure~\ref{fig:aaai-interface-summary}(b) splits 1000 validation examples by
whether the ranker-specific @5 window selects every annotated support
unit. The split is annotation-based and may miss alternative valid chains; the
support-first comparison is evaluated on the same subsets.

The split changes how the @5 score should be read: missing-support rows
measure a coverage failure, while complete-support rows ask whether the
shortened, ranked interface helps or hurts when annotated support survives.
This makes complete-support coverage a prerequisite for interpreting @5:
without it, the same answer score can point to a ranker problem, a reader problem,
or both.

\begin{table}[H]
\centering
\scriptsize
\setlength{\tabcolsep}{2.5pt}
\caption{Coverage split at 1000 examples (Qwen2.5-Coder-14B, train size 4000). Full means all support is selected; SF is a gold diagnostic. $*$: 95\% paired interval excludes zero; $\dagger$: HotpotQA missing bin ($n=52$).}
\label{tab:aaai-coverage-decomp}
\begin{tabular}{lllrcc}
\toprule
Dataset & Ranker & @5 support & $n$ & @5--raw & SF--raw \\
\midrule
HotpotQA & FT-CE & Full & 948 & +0.002 & +0.015 \\
HotpotQA & FT-CE & Missing$\dagger$ & 52 & -0.408* & +0.003 \\
2Wiki & CE & Full & 597 & +0.014 & +0.029* \\
2Wiki & CE & Missing & 403 & -0.280* & +0.062* \\
MuSiQue & CE & Full & 313 & +0.065* & +0.085* \\
MuSiQue & CE & Missing & 687 & -0.222* & +0.069* \\
\bottomrule
\end{tabular}
\end{table}

\paragraph{Coverage status separates severe @5 retrieval losses.}
When the ranker-specific @5 window selects all annotated support, it is tied with raw
on HotpotQA and 2Wiki and better than raw on MuSiQue. When support is missing,
the @5 window drops by 0.222--0.408 F1. The HotpotQA missing-support bin
contains 52 examples and has a strictly negative paired interval.
Aggregate top-$k$ F1 therefore mixes a reader-interface effect with a
support-coverage effect.

The same retrieval-window interface has different meanings across settings:
HotpotQA mostly tests compression with surviving support, 2Wiki mixes complete
and missing chains, and MuSiQue is coverage-limited under CE@5. Appendix C gives
the weighted bin breakdown and paired intervals.

\begin{table}[h]
\centering
\caption{Fixed-adapter support interventions on 1000 examples. Restore replaces
distractors with missing support; delete replaces one support unit. Deltas are
defined against the base @5 input; $*$ marks a 95\% paired interval excluding
zero. Appendix C gives intervals and the edit policy.}
\label{tab:aaai-support-intervention}
\scriptsize
\setlength{\tabcolsep}{2pt}
\begin{tabular}{lrrrrrr}
\toprule
Dataset & Missing $n$ & Base & Restore & Restore $\Delta$ & Full $n$ & Delete drop \\
\midrule
HotpotQA & 52 & 0.390 & 0.804 & +0.415* & 948 & +0.269* \\
2Wiki & 403 & 0.505 & 0.792 & +0.288* & 597 & +0.406* \\
MuSiQue & 687 & 0.326 & 0.610 & +0.284* & 313 & +0.337* \\
\bottomrule
\end{tabular}

\end{table}

\paragraph{A fixed-adapter intervention supports the coverage mechanism.}
Table~\ref{tab:aaai-support-intervention} edits the ranker-specific @5 window
after training while keeping the same adapter and window size. Restoring
missing support raises F1 by 0.284--0.415; deleting one selected support unit
drops F1 by 0.269--0.406. The paired intervals exclude zero on all datasets.
The same adapted @5 reader changes sharply when support
survival is changed directly.
Because the adapter is fixed, this intervention does not measure easier
training or a different learned output style; it measures whether the evaluated
input still contains the support the reader needs.

\paragraph{Support-first exposes remaining reader-interface effects.}
The support-first diagnostic is nonnegative in every coverage bin, although
some full-support gains are small and the HotpotQA missing bin is directional.
After coverage is separated, the reader-facing order and form can still matter.

MuSiQue shows the contrast most clearly: @5 is better than raw when the
annotated chain survives and much worse when it breaks. Appendix D gives
case-level examples.

\subsection{Ordering and Truncation Controls}

The support-first rows confound placement, grouping, and distractor order.
Table~\ref{tab:aaai-position-stress} therefore gives a small HotpotQA stress
test that keeps the raw candidate pool and varies annotated-support position.
Support-first is above shuffled, middle, and last placements, but the
intermediate rows are not monotonic. The result supports a weaker claim:
the complete support-first rendering helps, while position alone is not a
complete explanation.

\begin{table}[h]
\centering
\scriptsize
\setlength{\tabcolsep}{4pt}
\caption{HotpotQA position stress test with the 1B reader and 500 training examples.}
\label{tab:aaai-position-stress}
\begin{tabular}{lcc}
\toprule
Interface & F1 & Gain vs. raw \\
\midrule
Original raw & 0.379 & +0.000 \\
Support shuffled & 0.396 & +0.016 \\
Support middle & 0.401 & +0.021 \\
Support last & 0.443 & +0.063 \\
Support first & 0.539 & +0.159 \\
\bottomrule
\end{tabular}
\end{table}

Table~\ref{tab:aaai-order-controls} uses a Qwen2.5-Coder-14B train-size 2000
control setting. Ranked and @5 are the non-oracle selector rows in the main
paper: ranked keeps the full evidence pool but reorders it by the same ranker
used for the @5 row, while @5 then truncates that ranked pool.
The HotpotQA ranker is trained on training support labels and applied to
validation without validation gold labels; 2Wiki and MuSiQue use the
cross-encoder ranker. If ranked context and @5 diverge, ranking the pool
and truncating it must be considered separately.

\begin{table}[h]
\centering
\scriptsize
\setlength{\tabcolsep}{4.5pt}
\caption{Order and window controls with Qwen2.5-Coder-14B at train size 2000.
Ranked keeps all candidates under the same non-oracle ranker as @5; @5
truncates the ranked pool. SF uses gold support and is diagnostic.}
\label{tab:aaai-order-controls}
\begin{tabular}{llcccc}
\toprule
Dataset & Ranker & Raw & Ranked & @5 & SF$^\dagger$ \\
\midrule
HotpotQA & FT-CE & 0.748 & 0.754 & 0.755 & 0.801 \\
2Wiki & CE & 0.694 & 0.736 & 0.646 & 0.772 \\
MuSiQue & CE & 0.562 & 0.522 & 0.437 & 0.617 \\
\bottomrule
\end{tabular}
\end{table}

\paragraph{Reordering and truncation are different interventions.}
On HotpotQA, fine-tuned cross-encoder ranking and FT-CE@5 truncation are both slightly above raw in this
control setting. The expanded same-reader $n=1000$ BM25/CE grid (Appendix C)
shows full ranking tied with or above raw across datasets, while every @5 row
is lower. Thus ``using the retriever'' is not one intervention: ranking the
pool, truncating it, and oracle-moving support are distinct operations.

A weak lexical HotpotQA selector remains below raw context; the main HotpotQA
FT-CE already uses training support labels and reaches 0.948 CSR@5. Extending
support-supervised cross-encoders to the lower-coverage datasets raises CSR@5
from 0.597 to 0.959 on 2Wiki and from 0.313 to 0.688 on MuSiQue. With matched readers, FT-CE@5 reaches 0.794
versus 0.793 raw F1 and 0.596 versus 0.568, respectively, using shorter
prompts. Gold support-first remains higher at 0.835 and 0.642. Thus diagnostic
headroom is partly actionable but depends on selector coverage (Appendix C).

\subsection{Scaling and Generality}

The main matrix uses one large reader to keep the primary comparison clean;
Appendix C varies training-example count, reader family, adapter rank, seed, and
train--test interface pairing. Support-first stays positive, although its
HotpotQA magnitude shrinks for larger readers and training sets.

Three-seed Qwen2.5-Coder-14B gains range from +0.011 to +0.026 on HotpotQA,
+0.034 to +0.060 on 2Wiki, and +0.024 to +0.070 on MuSiQue. Cross-interface
evaluation is asymmetric: raw-trained adapters sometimes benefit from
support-first input, but support-first-trained adapters lose heavily on raw.
Thus the result is not merely an inference-time rewrite; the evaluation unit
is the adapted reader--interface pair.

A conventional, non-code-tuned Llama-3.1-8B replication evaluates raw, ranker@5, support-first,
coverage bins, and fixed-adapter interventions on 1000 examples per dataset.
The full-support @5--raw gaps are +0.016, +0.068, and +0.064, while the
missing-support gaps are -0.269, -0.113, and -0.179. Restore gains
(+0.208--+0.273) and delete drops (0.248--0.328) have paired intervals
excluding zero on all datasets (Appendix C).

\begin{table}[h]
\centering
\scriptsize
\setlength{\tabcolsep}{4pt}
\caption{Support-first F1 gains over raw.}
\label{tab:aaai-cross-reader-gains}
\begin{tabular}{lcccc}
\toprule
Dataset & Readers & Mean & Range & Positive \\
\midrule
HotpotQA & 6 & +0.046 & [+0.014, +0.081] & 6/6 \\
2Wiki & 7 & +0.075 & [+0.032, +0.115] & 7/7 \\
MuSiQue & 6 & +0.109 & [+0.070, +0.159] & 6/6 \\
\bottomrule
\end{tabular}
\end{table}

\paragraph{Support-first gains are not a single-reader artifact.}
Table~\ref{tab:aaai-cross-reader-gains} summarizes the available reader pool,
and Figure~\ref{fig:aaai-interface-summary}(c) shows the individual matched
gains. Every measured reader-dataset condition has a positive support-first gain
over raw context. This is directionality evidence, not a leaderboard, because
model families and train sizes differ. The full breakdown is in Appendix C.

\subsection{Evidence Reliance}

A remaining alternative is that adapters learn answer priors or output style
rather than using the evidence exposed by the interface. We test this by
removing annotated support after training.

\begin{table}[h]
\centering
\scriptsize
\setlength{\tabcolsep}{2.2pt}
\caption{Cross-dataset support removal with Qwen2.5-Coder-14B and BM25@5.
Cells are original-minus-removed F1; mean $\Delta$NLL is
removed-minus-original, averaged over the four interfaces.}
\label{tab:aaai-support-removal-multidataset}
\begin{tabular}{lccccc}
\toprule
Dataset & $\Delta$F1 Raw & $\Delta$F1 SF$^\dagger$ & $\Delta$F1 BM25@5 & $\Delta$F1 Oracle$^\dagger$ & Mean $\Delta$NLL \\
\midrule
HotpotQA & 0.535 & 0.615 & 0.404 & 0.540 & 1.815 \\
2Wiki & 0.487 & 0.566 & 0.352 & 0.549 & 0.998 \\
MuSiQue & 0.487 & 0.584 & 0.241 & 0.635 & 1.914 \\
\bottomrule
\end{tabular}
\end{table}

\paragraph{Removing support breaks the adapted readers.}
Table~\ref{tab:aaai-support-removal-multidataset} reports the same
support-removal diagnostic across all three datasets with Qwen2.5-Coder-14B.
The retrieval row uses BM25@5 to match the broader reliance checks.
F1 drops substantially after annotated
support is removed from raw context, support-first, BM25@5, and oracle support;
answer NLL rises for the same interfaces, arguing against an explanation based
only on answer priors, output postprocessing, or interface format. By contrast,
removing matched distractors changes F1 by at most 0.063 and often improves it,
making a generic deletion or length effect unlikely. HotpotQA 3B and
question-only checks support the same interpretation.

\section{Related Work}

\noindent\textbf{Retrieval and evidence selection.}
RAG systems improve reader evidence through retrieval, reranking, adaptive
generation, and reader-side training
\cite{karpukhin2020dpr,lewis2020rag,izacard2023atlas,shi2024replug,
asai2023selfrag,yu2024rankrag}. Sparse, dense, and cross-encoder methods rank
evidence differently \cite{chen2017drqa,robertson2009bm25,nogueira2019bert},
but the selected window is both retrieval outcome and reader input. Answer F1
therefore mixes chain loss with failure to use surviving support. Using
multi-hop annotations \cite{yang2018hotpotqa,ho2020twowiki,trivedi2022musique},
we condition on complete-chain survival and combine matched adaptation with
fixed-adapter evidence interventions. The contribution lies in this combined
protocol, which exposes when aggregate @5 averages opposite reader outcomes.

\noindent\textbf{Context use and adapted readers.}
Long-context evaluations document persistent difficulty in using long prompts,
while controlled probes show sensitivity to where evidence appears
\cite{bai2024longbench,liu2023lost}. Adapter and parameter-efficient tuning
shows that small trainable budgets adapt readers to new formats
\cite{hu2022lora,houlsby2019adapters,li2021prefixtuning}. Unlike inference-time
position probes, we retrain matched adapters per evidence form and show that
the top-$k$ effect can reverse after conditioning on complete-chain coverage.
Cross-interface asymmetry shows that adaptation is interface-specific, not
merely a position-sensitive inference-time response.
Gold-support rows remain diagnostic references, not serving proposals.

\section{Conclusion}

Across three support-annotated multi-hop QA datasets with closed candidate
pools, top-$k$ answer scores should be accompanied by complete-support coverage.
Gold support-first exposes reader-side headroom, and supervised
selectors show that part is actionable without closing the gold gap. The main
evaluation rule remains: check whether the chain survives before attributing
an answer change to the reader or retriever. Limitations are in Appendix~A.

\clearpage
\bibliography{cas-refs}

\clearpage
\appendix
\setcounter{secnumdepth}{1}
\numberwithin{table}{section}
\numberwithin{figure}{section}
\section{Discussion and Limitations}
\label{app:discussion-limitations}

The main paper treats evidence interfaces as a diagnostic object rather than a
new retrieval algorithm. This framing is useful because RAG failures often mix
two effects that are easy to confuse: the selected context may not contain the
full support chain, or it may contain that chain in a form that the adapted
reader uses poorly. The experiments therefore report answer quality together
with support coverage, ordering controls, scaling checks, and support-removal
diagnostics. The practical implication is an evaluation habit: before judging a
reader or a retriever from a single top-$k$ score, first ask whether the full
annotated support chain is selected, then ask whether the reader can use the surviving
evidence.

The study has several limits. The residual reader--interface effect is not treated as an independently
identified latent variable; it denotes the remaining difference
after annotated-support availability is measured. The oracle rows use annotated support or gold
structure, so they are diagnostic controls rather than deployable systems. A
weak lexical HotpotQA support-title selector remains below raw context; the
stronger support-supervised cross-encoder raises complete-chain coverage and
recovers part of the practical gap, but still leaves headroom to the gold
diagnostic and assumes support-labeled training data. The
datasets are multi-hop QA benchmarks with explicit support annotations, which
make the availability-usability split measurable but do not cover every RAG
setting; incomplete annotations or unmarked alternative support chains would
also affect the CSR bins. The matched LoRA protocol holds trainable capacity
fixed, but other adaptation regimes, larger instruction-tuned readers, mixed
interface formats, and tool-using agents may change the absolute scores. The
retrieval-window controls also combine several changes at once: prompt length,
ranking order, distractor removal, and possible support loss. The paper
therefore claims a practical decomposition, not a complete causal isolation of
every token-level factor.

Future work should replace the gold-support interfaces with learned interface
builders. A useful selector, compressor, or structure extractor should optimize
not only passage relevance or prompt length, but also preservation and
arrangement of the support chain for the target reader. The gaps reported in
the main paper give targets for such methods: the gap between raw context and
support-first shows reader-side headroom, while the gap between complete-support
and missing-support retrieval windows shows when retrieval coverage is still the
bottleneck. Extending the same diagnostics to open-ended generation, dialogue
RAG, and tasks without gold support labels would test whether the interface
view remains useful beyond controlled multi-hop QA.

\section{Experimental Details}

\subsection{Data Splits and Sampling}

All experiments are built from the official train and validation splits of
HotpotQA, 2WikiMultiHopQA, and MuSiQue. We subsample examples with a fixed
seed and keep the same selected source ids across interfaces within a
dataset-size condition. This matters because every interface comparison is
paired: raw context, support-first, top-$k$, oracle support, and structured
rows are evaluated on the same questions whenever the corresponding interface
exists.

The main Qwen2.5-Coder-14B matrix uses 4000 training examples and 300 validation
examples per dataset-interface row. Scaling and robustness tables use the
train sizes stated in their captions or table entries, but keep the same
principle: only the interface rendering changes inside a matched comparison.
The coverage-conditioned and fixed-adapter intervention analyses expand the
answer evaluation to 1000 matched validation examples per dataset.
During training, a small held-out subset is used only to monitor loss; the
reported answer metrics come from greedy validation generation.

\subsection{Evidence Units and Interface Rendering}

Each example contains a question $q_i$, an answer $a_i$, a candidate evidence
pool $C_i$, and, when available, annotated support $S_i$. For raw, ranked,
top-$k$, and support-first interfaces, HotpotQA and 2WikiMultiHopQA candidate
units are complete title-associated paragraphs. Their sentence-level
supporting facts identify support titles, but support-first reorders the whole
paragraph associated with each title; it does not extract supporting sentences.
MuSiQue candidate units are complete paragraphs and support-first likewise
moves whole annotated paragraphs. Oracle support is a separate compact
diagnostic: it extracts annotated supporting sentences for HotpotQA and 2Wiki,
while retaining annotated supporting paragraphs for MuSiQue. CSR checks title
coverage for HotpotQA/2Wiki and paragraph coverage for MuSiQue. We preserve
these dataset-native units rather than forcing a common segmentation.

Raw context preserves the dataset candidate order. Ranked context applies a
retrieval or reranking order but keeps the full candidate pool. Top-$k$ keeps
only the first $k$ ranked units. Support-first moves the complete support
paragraphs defined above before the remaining candidates while keeping
distractors in the input. Oracle support keeps only the compact annotated
support described above. Gold-chain triples are used
only for 2WikiMultiHopQA, where the annotated reasoning chain can be rendered
as entity--relation--answer triples. Because these triples often include the
answer string, this is an answer-exposing ceiling diagnostic, not a
deployable retrieval method.

All interfaces use the same answer prompt skeleton. Only the rendered evidence
block changes:

\begin{promptbox}
Question: [question text]

Evidence:
[ordered evidence units produced by the interface]

Answer:
\end{promptbox}

This template is intentionally plain. The study is about the evidence block,
not prompt engineering. Interface lengths are not forced to be equal because
length is part of the real design tradeoff between raw context, retrieval
windows, oracle support, and compact structure.

\subsection{Retrieval and Ordering Interfaces}

BM25 windows score each candidate unit against the question using lexical
matching and then keep either the full ranked order or the top-$k$ prefix.
Cross-encoder windows score each question--candidate pair independently using
the off-the-shelf \texttt{cross-encoder/ms-marco-MiniLM-L-6-v2} reranker and
sort candidates by the resulting score. For the main HotpotQA FT-CE row, the
checkpoint is fine-tuned on HotpotQA training support-title labels. The
supervised-ranker extension trains separate 2Wiki and MuSiQue copies on 5000
training examples from the corresponding dataset. Each question--candidate pair is labeled
positive if the candidate title appears in the training split's supporting
facts and negative otherwise. We train for one epoch with AdamW, learning rate
$2\cdot10^{-5}$, batch size 64, maximum length 384, 200 warmup steps, and the
training split's positive-to-negative class weight. The fitted ranker is then
applied to validation questions before the reader sees the selected window; no
validation support labels are used for ranking or tuning.

The position controls are designed to separate several effects that raw versus
support-first can otherwise mix. Support-last and support-middle keep the full
candidate pool but move annotated support to a later or middle region.
Support-shuffled randomizes support placement while preserving the full pool.
Random-first moves non-support material forward as a control for generic
reshuffling. Marked-in-place and grouped-in-place controls separate explicit
support marking and local grouping from front-loading. These controls use gold
support labels and are therefore diagnostic; they test what kind of interface
change matters, not what a deployed retriever already knows.

\subsection{Benchmark-Specific Support Interpretation}

HotpotQA uses a distractor setting with annotated supporting evidence
\cite{yang2018hotpotqa}. Complete-support recall asks whether the annotated
support units for the question appear inside the selected window.
2WikiMultiHopQA provides multi-hop reasoning paths and supports both text
support diagnostics and a compact gold-chain ceiling diagnostic
\cite{ho2020twowiki}. MuSiQue is built from composed single-hop questions and
is useful for testing whether short retrieval windows preserve longer evidence
chains \cite{trivedi2022musique}.

Support-unit granularity is dataset-specific. We therefore interpret support
recall within each dataset rather than comparing raw recall values as if the
annotations had identical granularity. The cross-dataset comparison is about
the diagnostic pattern: short windows can be cheap, but they must preserve all
support units needed by the reader.

\subsection{Reader Adaptation}

The adapted-reader experiments use LoRA updates on frozen causal language
models. For each interface, we train a separate adapter on the rendered
interface prompts and answer targets. The reader is not asked at test time to
switch among unseen formats. Instead, each interface defines a matched
adaptation problem under the same trainable parameter count.

Table~\ref{tab:supp-training-settings} lists the default settings. Unless a
table explicitly states otherwise, the main adapted-reader rows use rank-8
LoRA with alpha 16 and dropout 0.05 on the query, key, value, and output
projection matrices. We train for one epoch with AdamW, learning rate
\(2\times10^{-4}\), no weight decay, warmup ratio 0.03, max gradient norm 1.0,
and an effective batch size of 16 sequences. Full-context interfaces use a
larger sequence cap than short-window or oracle interfaces, matching the
natural length of the rendered input.

\begin{table}[h]
\centering
\scriptsize
\setlength{\tabcolsep}{3pt}
\renewcommand{\arraystretch}{1.05}
\caption{Default adaptation and generation settings. Table labels override these defaults when they report a different reader, train size, or diagnostic subset.}
\label{tab:supp-training-settings}
\begin{tabular}{@{}p{0.33\columnwidth}p{0.58\columnwidth}@{}}
\toprule
Setting & Value \\
\midrule
Main reader & Qwen2.5-Coder-14B-Instruct \\
Additional readers & Llama-3.2-1B/3B, Qwen2.5-Coder-7B, Mistral-7B, Llama-3.1-8B, Gemma-2-9B where reported \\
Adaptation & LoRA on a frozen causal reader \\
LoRA modules & query, key, value, and output projection matrices \\
Default LoRA config & rank 8, alpha 16, dropout 0.05 \\
Optimizer & AdamW, learning rate \(2\times10^{-4}\), weight decay 0, warmup ratio 0.03 \\
Training length & one epoch over the selected interface-rendered training set \\
Batching & effective batch size 16 via gradient accumulation \\
Train sizes & 4000 for main Qwen2.5-Coder-14B rows; smaller sizes are shown explicitly in scaling tables \\
Training length caps & 3072 tokens for full-context interfaces; 2048 for @5; 1536 for @3; shorter caps for oracle and no-context interfaces \\
Generation & greedy decoding, no sampling, at most 32 new tokens \\
Evaluation size & 300 validation examples per interface unless a table states otherwise \\
Uncertainty & paired bootstrap over matched validation source ids \\
\bottomrule
\end{tabular}
\end{table}

The main reader is Qwen2.5-Coder-14B-Instruct because it is the common 14B
checkpoint for which all three datasets and core interface families were
completed under the same protocol. The robustness tables include smaller Llama readers and several
7B--9B-class readers where available. Those rows are not used as a model
leaderboard; they test whether the support-first direction survives changes in
reader family and scale.
The released environment records Python 3.11 with PyTorch 2.5.1 (CUDA 12.1),
Transformers 5.9.0, PEFT 0.19.1, Accelerate 1.13.0, and
Sentence-Transformers 5.5.1.

\subsection{Generation, Metrics, and Uncertainty}

The main baselines are raw context, BM25 or cross-encoder top-$k$ retrieval
windows, and full ranked context. Support-first, oracle support, and
gold-chain triples are diagnostic controls that use dataset annotations when
available. At evaluation time, every adapted reader uses greedy generation
with no sampling and a 32-token output cap. We strip common answer prefixes
such as ``Answer:'' and score the first generated line.

We report exact match and token-level F1 after lowercasing, punctuation
removal, and article removal. For yes/no answers, containment checks require an
exact normalized yes/no match. Prompt-token cost is measured after applying the
model tokenizer and the chat template when the tokenizer provides one.
Complete-support recall is computed from the interface metadata by checking
whether every annotated support unit appears in the selected window. This is
an annotation-level selected-window proxy; the raw-context audit below checks
the stricter question of whether support text remains visible after final
tokenization and truncation.

Paired bootstrap intervals are computed over matched validation source ids.
This pairing is important because the same question can be rendered through
several interfaces; the uncertainty estimate therefore reflects the difference
between two interfaces on the same examples rather than variation from
different validation samples. We use 10,000 resamples and report percentile
95\% intervals with a fixed random seed (20260605 plus a deterministic
comparison offset). Support-removal diagnostics are evaluated after
training: annotated support units are removed from the validation input of a
trained adapter, and the resulting F1 and likelihood changes are compared with
the original interface evaluation. This tests whether the adapted reader is
actually relying on the exposed evidence rather than only learning an answer
style.

\section{Additional Experimental Results}

\subsection{Full Qwen2.5-Coder-14B Scaling}

Table~\ref{tab:supp-qwen14b-scaling} expands the scaling view summarized in
the main paper. Support-first remains above raw context at all measured train
sizes. The size of the gain changes by dataset: HotpotQA narrows as more
examples are available, while MuSiQue remains strongly positive. This pattern
is consistent with the main interpretation. More data can help a reader handle
raw context, but it does not remove the reader-side advantage of showing the
support chain in a cleaner order.

\begin{figure}[h]
\centering
\includegraphics[width=\columnwidth]{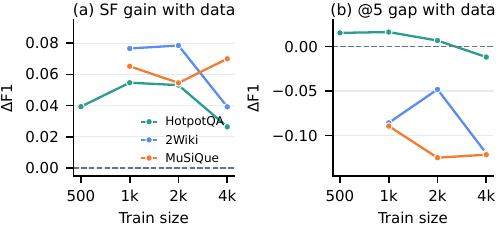}
\caption{Qwen2.5-Coder-14B scaling checks. Support-first gains remain positive as train
size grows, while ranker-specific @5 retrieval gaps stay dataset-dependent because they depend on
whether the retrieval window preserves the support chain.}
\label{fig:aaai-scaling-checks}
\end{figure}

\begin{table}[h]
\centering
\scriptsize
\setlength{\tabcolsep}{4pt}
\caption{Full Qwen2.5-Coder-14B train-size scaling by dataset and interface. Values are token-level F1.}
\label{tab:supp-qwen14b-scaling}
\resizebox{\columnwidth}{!}{%
\begin{tabular}{llcccc}
\toprule
Dataset & Interface & $n=500$ & $n=1000$ & $n=2000$ & $n=4000$ \\
\midrule
HotpotQA & Raw & 0.727 & 0.730 & 0.748 & 0.780 \\
 & SF & 0.766 & 0.785 & 0.801 & 0.806 \\
 & @5 & 0.742 & 0.746 & 0.755 & 0.768 \\
 & Oracle & 0.761 & 0.781 & 0.787 & 0.796 \\
\addlinespace
2Wiki & Raw & -- & 0.691 & 0.694 & 0.782 \\
 & SF & -- & 0.767 & 0.772 & 0.822 \\
 & @5 & -- & 0.605 & 0.646 & 0.663 \\
 & Oracle & -- & 0.789 & 0.816 & 0.844 \\
 & Gold tri. & -- & 0.993 & 0.993 & 0.993 \\
\addlinespace
MuSiQue & Raw & -- & 0.512 & 0.562 & 0.581 \\
 & SF & -- & 0.577 & 0.617 & 0.651 \\
 & @5 & -- & 0.422 & 0.437 & 0.459 \\
 & Oracle & -- & 0.689 & 0.692 & 0.747 \\
\bottomrule
\end{tabular}
}
\end{table}

The ranker-specific @5 window behaves differently from support-first. It remains close to raw on
HotpotQA but below raw on 2WikiMultiHopQA and MuSiQue. The full table therefore
separates two effects that are compressed in the main figure: support-first is
a reader-side ordering diagnostic, while the @5 window depends on whether a short
retrieval window preserves the complete chain.

\subsection{Cross-Interface Evaluation}

Table~\ref{tab:supp-cross-interface-eval} checks whether the support-first
effect is only an artifact of training a separate adapter for each interface.
The table reuses trained adapters and swaps the validation interface:
raw-trained adapters are evaluated on support-first inputs, and support-first
adapters are evaluated on raw inputs. The final row repeats the check for the
main Qwen2.5-Coder-14B settings.

\begin{table}[h]
\centering
\scriptsize
\setlength{\tabcolsep}{2.5pt}
\caption{Cross-interface evaluation. Rows report F1 when adapters trained on raw context or support-first (SF) are evaluated on raw or SF inputs.}
\label{tab:supp-cross-interface-eval}
\begin{tabular}{llccccc}
\toprule
Dataset & Reader & $n$ & Raw$\rightarrow$Raw & Raw$\rightarrow$SF & SF$\rightarrow$Raw & SF$\rightarrow$SF \\
\midrule
HotpotQA & 3B & 4000 & 0.719 & 0.721 & 0.427 & 0.733 \\
2Wiki & 3B & 4000 & 0.662 & 0.717 & 0.483 & 0.743 \\
MuSiQue & 3B & 4000 & 0.502 & 0.506 & 0.171 & 0.576 \\
HotpotQA & Qwen14B & 4000 & 0.780 & 0.806 & 0.659 & 0.806 \\
2Wiki & Qwen14B & 4000 & 0.782 & 0.809 & 0.637 & 0.822 \\
MuSiQue & Qwen14B & 4000 & 0.581 & 0.573 & 0.296 & 0.651 \\
\bottomrule
\end{tabular}
\end{table}

The pattern is asymmetric. A raw-trained adapter sometimes benefits from a
support-first input at evaluation time, especially on HotpotQA and 2Wiki, but
this alone does not consistently recover the fully matched support-first score.
In the reverse direction, support-first adapters lose heavily when evaluated on
raw context.
The result clarifies the claim boundary: the diagnostic measures how an
adapted reader learns and uses a consistent evidence interface, not a pure
inference-time prompt swap.

\subsection{No-Adapter Reader Control}

Table~\ref{tab:supp-frozen-reader-control} evaluates Qwen2.5-Coder-14B without
LoRA adaptation on the same 300 validation examples used by the main matrix.
The result gives a base-reader reference point for the adapted-reader
diagnostics. Support-first and oracle support are above raw context on all
three datasets, so the paper's effect is not created only by adapter training.
At the same time, this control is not the main object of measurement: the main
paper asks how a reader behaves after it is trained on a consistent evidence
form under a fixed adapter size and training-example count.

\begin{table}[h]
\centering
\scriptsize
\setlength{\tabcolsep}{3pt}
\caption{No-adapter Qwen2.5-Coder-14B control on 300 validation examples per dataset. The retrieval column is FT-CE@5 for HotpotQA and BM25@5 for 2Wiki and MuSiQue; daggered rows use gold annotations. Values are F1.}
\label{tab:supp-frozen-reader-control}
\begin{tabular}{lccccc}
\toprule
Dataset & Raw & Retr.@5 & SF$^\dagger$ & Oracle$^\dagger$ & Gold tri.$^\dagger$ \\
\midrule
HotpotQA & 0.630 & 0.626 & 0.665 & 0.738 & -- \\
2Wiki & 0.556 & 0.448 & 0.605 & 0.689 & 0.920 \\
MuSiQue & 0.353 & 0.244 & 0.450 & 0.557 & -- \\
\bottomrule
\end{tabular}
\end{table}

\subsection{Training-Seed Robustness}

Table~\ref{tab:supp-seed-robustness} repeats the raw versus support-first
comparison with two additional adapter-training seeds on HotpotQA, 2Wiki, and
MuSiQue. The validation examples, train size, model family, LoRA rank, and
decoding rule are held fixed; only the adapter initialization and training
order change. Seed 13 is the main matrix seed, and seeds 101 and 202 are
additional robustness runs.

\begin{table}[h]
\centering
\scriptsize
\setlength{\tabcolsep}{4pt}
\caption{Qwen2.5-Coder-14B seed robustness at train size 4000. Gains are F1 differences between support-first (SF) and raw context on the same validation slice.}
\label{tab:supp-seed-robustness}
\begin{tabular}{llccc}
\toprule
Dataset & Seed & Raw F1 & SF$^\dagger$ F1 & Gain \\
\midrule
HotpotQA & 13 & 0.780 & 0.806 & +0.026 \\
HotpotQA & 101 & 0.779 & 0.802 & +0.023 \\
HotpotQA & 202 & 0.791 & 0.802 & +0.011 \\
2Wiki & 13 & 0.782 & 0.822 & +0.040 \\
2Wiki & 101 & 0.758 & 0.792 & +0.034 \\
2Wiki & 202 & 0.757 & 0.817 & +0.060 \\
MuSiQue & 13 & 0.581 & 0.651 & +0.070 \\
MuSiQue & 101 & 0.593 & 0.652 & +0.059 \\
MuSiQue & 202 & 0.613 & 0.637 & +0.024 \\
\bottomrule
\end{tabular}
\end{table}

The support-first gain remains positive in all nine matched rows. The exact
magnitude varies: HotpotQA is consistently modest, 2Wiki remains larger, and
MuSiQue is positive but more variable across seeds. We therefore do not treat
the seed table as a new scaling law. Its role is narrower: the support-first
direction in the main matrix is not explained by a single favorable
adapter-training seed.

\subsection{Question-Only Diagnostic}

Table~\ref{tab:supp-no-context} reports the question-only diagnostic at train
size 2000. No-context F1 is far below evidence-bearing interfaces on all three
datasets. This does not prove that the reader never uses parametric knowledge,
but it rules out the simple explanation that the main results are driven only
by answer priors.

\begin{table}[h]
\centering
\scriptsize
\setlength{\tabcolsep}{3pt}
\caption{Qwen2.5-Coder-14B no-context diagnostic at train size 2000. Values are F1.}
\label{tab:supp-no-context}
\begin{tabular}{lrrrrrr}
\toprule
Dataset & No ctx. & Raw & @5 & SF & Oracle & Gold tri. \\
\midrule
HotpotQA & 0.280 & 0.748 & 0.755 & 0.801 & 0.787 & -- \\
2Wiki & 0.315 & 0.694 & 0.646 & 0.772 & 0.816 & 0.993 \\
MuSiQue & 0.144 & 0.562 & 0.437 & 0.617 & 0.692 & -- \\
\bottomrule
\end{tabular}
\end{table}

The gap is especially important for the support-removal analysis. If the
adapted readers were mostly memorizing answer styles or exploiting dataset
priors, removing support would not cause such large drops. The no-context rows
show that evidence-free answering is weak in the same benchmark setting.

\subsection{Raw-Context Truncation Audit}

Table~\ref{tab:supp-raw-truncation} checks a simpler reader-input issue: when
raw context is rendered with the Qwen2.5-Coder chat template and capped at
3072 input tokens, does annotated support survive the actual model input? This
audit does not train or evaluate a reader. It only compares the stored
interface text with the truncated prompt that the reader would receive.

\begin{table}[h]
\centering
\caption{Raw-context truncation audit with the Qwen2.5-Coder tokenizer and the
3072-token reader budget. Title survival is computed where support-title
metadata is available; support-text survival uses unit-level matching against
the rendered gold support.}
\label{tab:supp-raw-truncation}
\small
\setlength{\tabcolsep}{3pt}
\begin{tabular}{lrrrr}
\toprule
Dataset & $n$ & Trunc. & Title surv. & Text surv. \\
\midrule
HotpotQA & 1000 & 1 & 1/1 & 1/1 \\
2Wiki & 1000 & 2 & 2/2 & 2/2 \\
MuSiQue & 1000 & 140 & 123/140 & 91/140 \\
\bottomrule
\end{tabular}
\end{table}

The audit separates HotpotQA and 2Wiki from MuSiQue. Almost no HotpotQA or
2Wiki raw prompts exceed the 3072-token budget in the evaluated validation
slice. MuSiQue is different: 140 of 1000 raw prompts are truncated, and within
those truncated prompts only 123 retain every support title and 91 retain every
matched support-text unit under the unit-level matching heuristic. This does
not explain all MuSiQue failures, but it gives a concrete input-side reason
why long raw context and short retrieval windows need to be interpreted
together with support-survival diagnostics.
The title and text columns are not interchangeable: title survival is a
metadata-level check, while text survival uses conservative exact matching
against rendered support units. Exact text matching can miss cases where the
benchmark annotation, rendered paragraph boundary, or normalization differs
from the final prompt even though the title remains present.

This audit raises a stricter check for the raw-versus-support-first comparison:
the gain should not be an artifact of raw prompts losing annotated support after
tokenization and truncation. Table~\ref{tab:supp-visible-support-audit}
extends the audit to the main 300-example slice and includes raw context,
support-first, and the ranker-specific @5 window. No @5 prompt is truncated. The
text-visible counts track final support visibility under a conservative exact
unit-level heuristic: title visibility can overestimate support survival when a
title appears but the annotated support text does not. MuSiQue shows the
clearest input-level interface effect. Raw and support-first have the same
length, but support-first places support before the token cap, so all MuSiQue
support text remains visible in the final prompt.

\begin{table}[h]
\centering
\scriptsize
\setlength{\tabcolsep}{3pt}
\caption{Final-tokenized support visibility in the main 300-example Qwen2.5-Coder-14B validation slice. Prompts are rendered with the chat template and capped at 3072 input tokens. Title visibility is reported where support titles are available; text visibility uses exact unit-level matching against rendered gold support.}
\label{tab:supp-visible-support-audit}
\resizebox{\columnwidth}{!}{%
\begin{tabular}{llrrrr}
\toprule
Dataset & Interface & Trunc. & Avg tok. & Title visible & Text visible \\
\midrule
HotpotQA & Raw & 1 & 1485 & 300/300 & 295/300 \\
HotpotQA & SF & 1 & 1485 & 300/300 & 295/300 \\
HotpotQA & @5 & 0 & 685 & 290/300 & 281/300 \\
2Wiki & Raw & 0 & 1059 & 300/300 & 282/300 \\
2Wiki & SF & 0 & 1059 & 300/300 & 282/300 \\
2Wiki & @5 & 0 & 624 & 287/300 & 163/300 \\
MuSiQue & Raw & 36 & 2523 & 295/300 & 282/300 \\
MuSiQue & SF & 36 & 2523 & 300/300 & 300/300 \\
MuSiQue & @5 & 0 & 700 & 153/300 & 106/300 \\
\bottomrule
\end{tabular}
}
\end{table}

Table~\ref{tab:supp-visible-support-filter} then filters the main 300-example
slice to examples where annotated support text survives in both final rendered
raw and support-first prompts. The support-first direction remains positive on
all three datasets, including MuSiQue, where the filtered gain is +0.075 F1.
Raw truncation is still part of the interface being diagnosed, but it does not
explain away the support-first effect.

\begin{table}[h]
\centering
\caption{Raw versus support-first after filtering to examples where annotated
support text survives in both final model-visible prompts. Visibility is checked
after applying the Qwen2.5-Coder chat template and input-token cap. Values are
F1 on the main 300-example evaluation slice.}
\label{tab:supp-visible-support-filter}
\small
\setlength{\tabcolsep}{4pt}
\begin{tabular}{lrrrr}
\toprule
Dataset & Visible $n$ & Raw & SF$^\dagger$ & Gain \\
\midrule
HotpotQA & 295 & 0.779 & 0.804 & +0.025 \\
2Wiki & 282 & 0.792 & 0.819 & +0.027 \\
MuSiQue & 282 & 0.616 & 0.692 & +0.075 \\
\bottomrule
\end{tabular}
\end{table}

\subsection{Retrieval Support Recall}

Table~\ref{tab:supp-retrieval-recall} gives the full retrieval support-recall
view behind the main retrieval-window analysis. The main paper focuses on the
ranker used in the @5 answer-quality comparison, while this table also
shows BM25 and alternate windows.

\begin{table}[h]
\centering
\scriptsize
\setlength{\tabcolsep}{3pt}
\caption{Full retrieval support recall for the evaluated rankers and window sizes. All support is complete-support recall; mean fraction is partial-chain coverage.}
\label{tab:supp-retrieval-recall}
\resizebox{\columnwidth}{!}{%
\begin{tabular}{llrrrr}
\toprule
Dataset & Ranker & $k$ & Any support & All support & Mean fraction \\
\midrule
HotpotQA & BM25 & 5 & 0.994 & 0.702 & 0.848 \\
HotpotQA & BM25 & 10 & 1.000 & 1.000 & 1.000 \\
HotpotQA & BM25 & 20 & 1.000 & 1.000 & 1.000 \\
HotpotQA & Cross-encoder & 5 & 0.995 & 0.764 & 0.879 \\
HotpotQA & Cross-encoder & 10 & 1.000 & 1.000 & 1.000 \\
HotpotQA & Cross-encoder & 20 & 1.000 & 1.000 & 1.000 \\
HotpotQA & Fine-tuned cross-encoder & 5 & 0.999 & 0.948 & 0.974 \\
HotpotQA & Fine-tuned cross-encoder & 10 & 1.000 & 1.000 & 1.000 \\
HotpotQA & Fine-tuned cross-encoder & 20 & 1.000 & 1.000 & 1.000 \\
2Wiki & BM25 & 5 & 0.999 & 0.534 & 0.789 \\
2Wiki & BM25 & 10 & 1.000 & 1.000 & 1.000 \\
2Wiki & BM25 & 20 & 1.000 & 1.000 & 1.000 \\
2Wiki & Cross-encoder & 5 & 1.000 & 0.597 & 0.824 \\
2Wiki & Cross-encoder & 10 & 1.000 & 1.000 & 1.000 \\
2Wiki & Cross-encoder & 20 & 1.000 & 1.000 & 1.000 \\
MuSiQue & BM25 & 5 & 0.931 & 0.206 & 0.571 \\
MuSiQue & BM25 & 10 & 0.975 & 0.412 & 0.717 \\
MuSiQue & BM25 & 20 & 1.000 & 1.000 & 1.000 \\
MuSiQue & Cross-encoder & 5 & 0.966 & 0.313 & 0.651 \\
MuSiQue & Cross-encoder & 10 & 0.993 & 0.527 & 0.786 \\
MuSiQue & Cross-encoder & 20 & 1.000 & 1.000 & 1.000 \\
\bottomrule
\end{tabular}
}
\end{table}

The table explains why aggregate top-$k$ F1 is hard to interpret. HotpotQA
reaches complete support quickly under the fine-tuned cross-encoder, so the FT-CE@5 window
mostly acts as compression. MuSiQue needs much larger windows to reach the same
coverage, so a short top-$k$ window is often missing part of the chain. The
``any support'' column is easier than the complete-support column, which
matters because partial evidence can still leave the reader without the hop
needed for the final answer.

Table~\ref{tab:supp-uniform-ranker-grid} removes a remaining comparison
asymmetry by evaluating raw context, full BM25 and cross-encoder orderings,
their @5 truncations, and gold support-first with the same reader family,
adapter rank, training size, and 1000-example evaluation budget. Full ordering
and truncation are reported separately because a ranker can improve the order
of the candidate pool while its short window still drops part of the support
chain.

\begin{table}[h]
\centering
\scriptsize
\setlength{\tabcolsep}{2.5pt}
\caption{Uniform ranker grid with Qwen2.5-Coder-14B (train size 2000; eval $n=1000$). Full keeps all ranked candidates; @5 cells show F1 with CSR in parentheses. SF is the gold support-first diagnostic.}
\label{tab:supp-uniform-ranker-grid}
\resizebox{\columnwidth}{!}{%
\begin{tabular}{lrrrrrr}
\toprule
Dataset & Raw & BM25 full & BM25@5 & CE full & CE@5 & SF \\
\midrule
HotpotQA & 0.756 & 0.757 & 0.673 (0.70) & 0.773 & 0.680 (0.76) & 0.795 \\
2Wiki & 0.722 & 0.770 & 0.643 (0.53) & 0.778 & 0.672 (0.60) & 0.786 \\
MuSiQue & 0.535 & 0.550 & 0.360 (0.21) & 0.536 & 0.424 (0.31) & 0.593 \\
\bottomrule
\end{tabular}
}
\end{table}

Across all six dataset--ranker pairs, full ordering is tied with or above raw
context, whereas every @5 truncation is lower. The largest @5 losses occur at
the lowest CSR values: MuSiQue BM25@5 has 0.21 CSR and $-0.175$ F1, while
CE@5 has 0.31 CSR and $-0.111$ F1. Thus the grid attributes the main loss to
short-window support omission rather than to ranking the full pool.

\subsection{Weighted Coverage Decomposition}

Table~\ref{tab:supp-weighted-coverage} converts the coverage-conditioned rows
from the main paper into a dataset-level decomposition. The table asks how much
of the aggregate @5--raw gap comes from examples whose support chain is
complete in the window, and how much comes from examples where at least one
annotated support unit is missing. This is a bookkeeping analysis rather than a
new model run, but it makes the interpretation of the ranker-specific @5 rows more precise.

\begin{table}[h]
\centering
\scriptsize
\setlength{\tabcolsep}{3pt}
\caption{Weighted 1000-example @5 coverage decomposition. If all complete is the counterfactual @5--raw gap if missing-support examples behaved like complete-support examples; missing extra is the additional weighted penalty from the missing-support bin.}
\label{tab:supp-weighted-coverage}
\resizebox{\columnwidth}{!}{%
\begin{tabular}{lrrrrrr}
\toprule
Dataset & Missing & Full gap & Missing gap & Agg. & If full & Missing extra \\
\midrule
HotpotQA & 5.2\% & +0.002 & -0.408 & -0.019 & +0.002 & -0.021 \\
2Wiki & 40.3\% & +0.014 & -0.280 & -0.104 & +0.014 & -0.118 \\
MuSiQue & 68.7\% & +0.065 & -0.222 & -0.132 & +0.065 & -0.197 \\
\bottomrule
\end{tabular}
}
\end{table}

The decomposition separates three cases. HotpotQA has little missing support,
so the aggregate @5 gap is small even though the missing-support subset is
bad. 2Wiki has a mixed window: the complete-support bin is nearly tied with
raw, but the missing-support bin is large enough to account for almost all of
the aggregate loss. MuSiQue is the clearest warning case. When the full chain
survives, @5 is above raw; after weighting by the much larger
missing-support share, the aggregate @5 row becomes negative. The main
paper therefore reports coverage-conditioned rows instead of treating top-$k$
F1 as a single reader-quality number.

\subsection{Coverage and Intervention Uncertainty}

Table~\ref{tab:supp-coverage-ci} reports paired bootstrap intervals for the
coverage-conditioned rows in the main paper. The intervals preserve the
matched-example structure: each resample draws validation examples within the
same support-status bin and recomputes the paired difference against raw
context. The expanded evaluation raises the HotpotQA missing-support bin from
16 to 52 examples and makes its @5--raw interval strictly negative. The 2Wiki
and MuSiQue missing-support bins are also strictly negative, while the MuSiQue complete-support bin is
strictly positive. This is the central ambiguity of aggregate @5 F1: a short
window can help when the chain survives and hurt when it drops a support unit.

\begin{table}[h]
\centering
\scriptsize
\setlength{\tabcolsep}{2.2pt}
\caption{Paired bootstrap intervals for the 1000-example coverage decomposition. Intervals resample matched examples within each support-status bin.}
\label{tab:supp-coverage-ci}
\resizebox{\columnwidth}{!}{%
\begin{tabular}{llrcc}
\toprule
Dataset & Bin & $n$ & @5--raw & SF--raw \\
\midrule
HotpotQA & Full & 948 & +0.002 [-0.016, +0.020] & +0.015 [-0.002, +0.032] \\
HotpotQA & Missing & 52 & -0.408 [-0.538, -0.276] & +0.003 [-0.079, +0.082] \\
2Wiki & Full & 597 & +0.014 [-0.009, +0.038] & +0.029 [+0.010, +0.049] \\
2Wiki & Missing & 403 & -0.280 [-0.330, -0.230] & +0.062 [+0.031, +0.094] \\
MuSiQue & Full & 313 & +0.065 [+0.025, +0.106] & +0.085 [+0.052, +0.121] \\
MuSiQue & Missing & 687 & -0.222 [-0.264, -0.180] & +0.069 [+0.042, +0.095] \\
\bottomrule
\end{tabular}
}
\end{table}

Table~\ref{tab:supp-intervention-ci} gives the same uncertainty view for the
fixed-adapter support intervention. Restore is evaluated only on rows where
the ranker-specific @5 window misses annotated support; delete is evaluated only on
rows where the @5 window selects complete annotated support. Restore replaces
non-support documents in the five-document window with the missing support
documents from the raw candidate pool, choosing replaceable documents with
similar rough word length where possible. Delete removes the latest support
document in the displayed @5 window and replaces it with a non-window,
non-support distractor of closest rough word length. Both interventions
preserve the question, prompt template, trained adapter, and five-document
window size. The restore intervals and delete-drop intervals are positive
across datasets. This supports the coverage mechanism without retraining the
reader: the same adapter changes answer quality when the visible support chain
is directly changed.

\begin{table}[h]
\centering
\scriptsize
\setlength{\tabcolsep}{3pt}
\caption{Paired bootstrap intervals for fixed-adapter support interventions on the 1000-example evaluation. Restore is evaluated on missing-support @5 rows; delete is evaluated on full-support @5 rows.}
\label{tab:supp-intervention-ci}
\resizebox{\columnwidth}{!}{%
\begin{tabular}{lrrrr}
\toprule
Dataset & Missing $n$ & Restore $\Delta$ & Full $n$ & Delete drop \\
\midrule
HotpotQA & 52 & +0.415 [+0.287, +0.544] & 948 & +0.269 [+0.240, +0.298] \\
2Wiki & 403 & +0.288 [+0.242, +0.333] & 597 & +0.406 [+0.366, +0.445] \\
MuSiQue & 687 & +0.284 [+0.249, +0.319] & 313 & +0.337 [+0.289, +0.387] \\
\bottomrule
\end{tabular}
}
\end{table}

\subsection{Top-10 Retrieval Windows}

Table~\ref{tab:supp-top10-windows} reports additional top-10 retrieval-window
controls. These rows address a specific interpretation risk: if only the @5
windows are shown, a reader may conclude that the non-oracle baseline is
artificially weak. The top-10 rows test whether a larger realistic window
closes the gap before any gold-support diagnostic is used.

\begin{table}[h]
\centering
\scriptsize
\setlength{\tabcolsep}{4pt}
\caption{Top-10 retrieval-window checks. Qwen2.5-Coder-14B rows use BM25 at train size 2000; 3B BM25 rows use train size 4000; CE rows use train size 2000.}
\label{tab:supp-top10-windows}
\begin{tabular}{llrrrr}
\toprule
Dataset & Reader/window & EM & F1 & Ans. & Tokens \\
\midrule
HotpotQA & Qwen2.5-Coder-14B & 0.623 & 0.773 & 0.667 & 1479 \\
2Wiki & Qwen2.5-Coder-14B & 0.630 & 0.696 & 0.673 & 1046 \\
MuSiQue & Qwen2.5-Coder-14B & 0.360 & 0.472 & 0.383 & 1318 \\
HotpotQA & 3B & 0.607 & 0.738 & 0.637 & 1423 \\
2Wiki & 3B & 0.650 & 0.713 & 0.687 & 995 \\
MuSiQue & 3B & 0.310 & 0.411 & 0.317 & 1267 \\
HotpotQA & 3B CE & 0.583 & 0.715 & 0.620 & 1423 \\
HotpotQA & 3B FT-CE & 0.587 & 0.718 & 0.610 & 1423 \\
2Wiki & 3B CE & 0.587 & 0.662 & 0.627 & 998 \\
MuSiQue & 3B CE & 0.323 & 0.416 & 0.340 & 1281 \\
\bottomrule
\end{tabular}
\end{table}

The answer is mixed. On HotpotQA and 2Wiki, BM25@10 recovers much of the loss
seen with BM25@5, and the 3B rows show that top-10 can be competitive when the
annotated support chain is selected. MuSiQue remains lower even with top-10, which is
consistent with the support-recall table: longer composed chains make short
and medium windows more fragile. The cross-encoder rows are not presented as a
new reranker benchmark. They provide an additional non-oracle check that
better ranking changes the window but does not remove the need to measure
annotated-support coverage.

\subsection{Non-Oracle Support Ordering}

Table~\ref{tab:supp-learned-support-ordering} adds a stricter non-oracle
support-ordering check on HotpotQA. The support-title selector is a TF-IDF
logistic model trained only on training support titles and then applied to
validation examples without validation support labels. It tests whether the
support-first target can be approximated by a learned selector rather than by
gold validation annotations. The answer is only partial: full ordering reaches
0.747 F1 and its @5 window reaches 0.647. The stronger FT-CE ranker is closer to
raw context, but still below the gold support-first diagnostic. This result is a
useful boundary condition for the paper's claim: support-first is a diagnostic
target, and turning it into a deployable interface requires a sufficiently
accurate selector.

\begin{table}[h]
\centering
\scriptsize
\setlength{\tabcolsep}{3.4pt}
\caption{HotpotQA non-oracle support-ordering checks with Qwen2.5-Coder-14B at train size 4000. The support-title selector is trained on training support titles and applied to validation examples without validation gold labels. FT-CE is the stronger learned ranker used in the main @5 controls.}
\label{tab:supp-learned-support-ordering}
\begin{tabular}{lcccc}
\toprule
Interface & EM & F1 & Ans. cont. & Avg. tok. \\
\midrule
Raw context & 0.627 & 0.780 & 0.677 & 1484 \\
Support-first$^\dagger$ & 0.657 & 0.806 & 0.690 & 1484 \\
FT-CE order & 0.630 & 0.782 & 0.677 & 1484 \\
FT-CE@5 & 0.623 & 0.768 & 0.670 & 685 \\
Support-title order & 0.600 & 0.747 & 0.660 & 1484 \\
Support-title@5 & 0.507 & 0.647 & 0.553 & 630 \\
\bottomrule
\end{tabular}
\end{table}

A stronger support-supervised cross-encoder provides a second, cross-dataset
test of this boundary. It is trained only on training-split support titles or
paragraph labels, then used to construct @5 inputs for separately matched
readers. Table~\ref{tab:supp-supervised-ranker} reports both answer F1 and
complete-support recall, together with coverage-conditioned gaps against raw
context. This distinguishes failure of the learned selector from failure of a
reader given a complete selected chain.

\begin{table}[h]
\centering
\scriptsize
\setlength{\tabcolsep}{2.5pt}
\caption{Support-supervised FT-CE@5 on Qwen2.5-Coder-14B (ranker train size 5000; reader train size 4000; eval $n=1000$). Parentheses give CSR. SF is the gold diagnostic.}
\label{tab:supp-supervised-ranker}
\resizebox{\columnwidth}{!}{%
\begin{tabular}{lrrrrrr}
\toprule
Dataset & Raw & CE@5 & FT-CE@5 & SF & Full gap & Missing gap \\
\midrule
2Wiki & 0.793 & 0.689 (0.60) & 0.794 (0.96) & 0.835 & +0.018 & -0.395 \\
MuSiQue & 0.568 & 0.436 (0.31) & 0.596 (0.69) & 0.642 & +0.121 & -0.179 \\
\bottomrule
\end{tabular}
}
\end{table}

The support-supervised ranker raises CSR@5 from 0.597 to 0.959 on 2Wiki and
from 0.313 to 0.688 on MuSiQue. Its matched-reader F1 is 0.794 on 2Wiki,
essentially tied with 0.793 raw context, and 0.596 on MuSiQue, above 0.568 raw
context. Mean prompt tokens fall from 1055 to 741 and from 2465 to 772,
respectively. The overall paired FT-CE@5--raw gaps are +0.001
[-0.019,+0.021] and +0.028 [-0.001,+0.056], so we describe the point estimates
as matching or recovering raw quality rather than as significant improvements.
The remaining gold support-first headroom is 0.041 and 0.046 F1.
Coverage conditioning preserves the paper's mechanism: FT-CE@5 is positive
against raw when the full chain survives and sharply negative when it does not.

\subsection{Adapter Capacity}

\begin{figure}[h]
\centering
\includegraphics[width=\columnwidth]{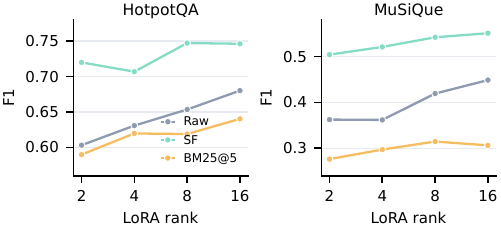}
\caption{LoRA-rank sensitivity with the 3B reader at train size 2000. Higher
rank improves some absolute scores, but support-first remains above raw and
BM25@5 across the measured ranks.}
\label{fig:aaai-rank-capacity}
\end{figure}

Figure~\ref{fig:aaai-rank-capacity} and Table~\ref{tab:supp-rank-sweep} give
the LoRA-rank sweep. The sweep uses the 3B reader at train size 2000 and varies
only the LoRA rank while keeping the interface and evaluation split fixed.

\begin{table}[h]
\centering
\scriptsize
\setlength{\tabcolsep}{4pt}
\caption{LoRA rank sweep with the 3B reader at train size 2000. Values are F1.}
\label{tab:supp-rank-sweep}
\begin{tabular}{llrrrr}
\toprule
Dataset & Interface & $r=2$ & $r=4$ & $r=8$ & $r=16$ \\
\midrule
HotpotQA & Raw & 0.603 & 0.631 & 0.653 & 0.680 \\
HotpotQA & SF & 0.720 & 0.707 & 0.747 & 0.746 \\
HotpotQA & BM25@5 & 0.590 & 0.620 & 0.619 & 0.640 \\
MuSiQue & Raw & 0.362 & 0.362 & 0.419 & 0.449 \\
MuSiQue & SF & 0.504 & 0.521 & 0.542 & 0.551 \\
MuSiQue & BM25@5 & 0.276 & 0.297 & 0.315 & 0.306 \\
\bottomrule
\end{tabular}
\end{table}

Increasing rank improves some absolute scores, especially for raw context, but
the support-first ordering remains above raw and BM25@5 across the measured
ranks. This is the relevant robustness point. The paper does not claim that
rank 16 is optimal, or that LoRA rank effects are identical across model
families. The sweep only rules out a narrow artifact explanation: the
support-first advantage is not produced by one unlucky rank-8 capacity choice.

\subsection{Full Support-Removal Diagnostics}

Table~\ref{tab:supp-reliance-full} expands the support-removal diagnostic
summarized in the main paper. Each row evaluates a trained adapter twice: once
with the original validation interface and once after annotated support is
removed from the same interface. The comparison is therefore not a new
training run; it tests whether the trained reader depends on the evidence
exposed by the interface at evaluation time.

\begin{table}[h]
\centering
\scriptsize
\setlength{\tabcolsep}{3pt}
\caption{Full support-removal diagnostics. F1 w/o $S$ removes annotated support from the same trained reader input; lower NLL is better.}
\label{tab:supp-reliance-full}
\resizebox{\columnwidth}{!}{%
\begin{tabular}{lllrrrrr}
\toprule
Dataset & Reader & Interface & F1 & F1 w/o $S$ & Drop & NLL & NLL w/o $S$ \\
\midrule
HotpotQA & Qwen2.5-Coder-14B & Raw & 0.780 & 0.244 & 0.535 & 0.347 & 2.407 \\
HotpotQA & Qwen2.5-Coder-14B & SF & 0.806 & 0.191 & 0.615 & 0.315 & 2.628 \\
HotpotQA & Qwen2.5-Coder-14B & BM25@5 & 0.690 & 0.287 & 0.404 & 0.613 & 1.873 \\
HotpotQA & Qwen2.5-Coder-14B & Oracle & 0.796 & 0.256 & 0.540 & 0.293 & 1.919 \\
HotpotQA & 3B & Raw & 0.719 & 0.240 & 0.479 & 0.391 & 2.582 \\
HotpotQA & 3B & SF & 0.741 & 0.156 & 0.585 & 0.388 & 3.132 \\
HotpotQA & 3B & Oracle & 0.773 & 0.259 & 0.515 & 0.338 & 2.019 \\
2Wiki & Qwen2.5-Coder-14B & Raw & 0.782 & 0.296 & 0.487 & 0.218 & 1.364 \\
2Wiki & Qwen2.5-Coder-14B & SF & 0.822 & 0.255 & 0.566 & 0.187 & 1.555 \\
2Wiki & Qwen2.5-Coder-14B & BM25@5 & 0.629 & 0.276 & 0.352 & 0.582 & 1.116 \\
2Wiki & Qwen2.5-Coder-14B & Oracle & 0.844 & 0.295 & 0.549 & 0.148 & 1.092 \\
2Wiki & Qwen2.5-Coder-14B & Gold tri. & 0.993 & 0.322 & 0.672 & 0.007 & 1.150 \\
2Wiki & 3B & Raw & 0.666 & 0.274 & 0.392 & 0.311 & 1.483 \\
2Wiki & 3B & SF & 0.745 & 0.211 & 0.534 & 0.256 & 1.867 \\
2Wiki & 3B & BM25@5 & 0.582 & 0.290 & 0.292 & 0.589 & 1.168 \\
2Wiki & 3B & Oracle & 0.823 & 0.307 & 0.516 & 0.180 & 1.093 \\
2Wiki & 3B & Gold tri. & 1.000 & 0.281 & 0.718 & 0.002 & 1.197 \\
MuSiQue & Qwen2.5-Coder-14B & Raw & 0.581 & 0.093 & 0.487 & 0.396 & 2.774 \\
MuSiQue & Qwen2.5-Coder-14B & SF & 0.651 & 0.067 & 0.584 & 0.307 & 3.006 \\
MuSiQue & Qwen2.5-Coder-14B & BM25@5 & 0.377 & 0.136 & 0.241 & 1.177 & 1.985 \\
MuSiQue & Qwen2.5-Coder-14B & Oracle & 0.747 & 0.112 & 0.635 & 0.281 & 2.055 \\
MuSiQue & 3B & Raw & 0.502 & 0.099 & 0.403 & 0.552 & 3.108 \\
MuSiQue & 3B & SF & 0.579 & 0.053 & 0.527 & 0.401 & 3.447 \\
MuSiQue & 3B & BM25@5 & 0.320 & 0.118 & 0.202 & 1.397 & 2.285 \\
MuSiQue & 3B & Oracle & 0.657 & 0.119 & 0.538 & 0.381 & 2.360 \\
\bottomrule
\end{tabular}
}
\end{table}

The same pattern appears across datasets and readers. Removing support causes
large F1 drops and increases token-weighted answer NLL for raw context,
support-first, BM25@5, oracle support, and the 2Wiki structured diagnostic.
The drops are often largest for support-first and oracle rows, which is
expected: those interfaces make the annotated support easier to use, so
removing it removes the reader's main signal. This strengthens the evidence
reliance claim in the main paper and makes it less dependent on a single
HotpotQA control.

\begin{figure}[h]
\centering
\includegraphics[width=\columnwidth]{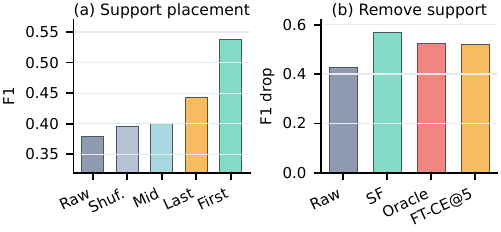}
\caption{Reader-side diagnostics. (a) The gold support-first rendering improves
HotpotQA F1 in a fixed raw-evidence pool; position-only controls are not
monotonic. (b) Removing annotated support after training produces large F1
drops across interfaces.}
\label{fig:aaai-diagnostic-checks}
\end{figure}

\subsection{Matched Distractor-Removal Control}

Table~\ref{tab:supp-matched-distractor} checks whether the support-removal
effect is mainly caused by shortening or perturbing the input. For each
validation example, we remove non-support text blocks whose lengths are matched
to the annotated support blocks when such distractors are available. The
reader, adapter, decoding rule, validation examples, and interface family are
otherwise unchanged.

\begin{table}[h]
\centering
\scriptsize
\setlength{\tabcolsep}{4.0pt}
\caption{Matched distractor-removal control for the Qwen2.5-Coder-14B support-removal diagnostic. $\Delta$F1 is original F1 minus ablated F1; negative values mean the ablated input improved.}
\label{tab:supp-matched-distractor}
\begin{tabular}{llrrrr}
\toprule
Dataset & Interface & Orig. F1 & F1 w/o $S$ & $\Delta$F1 w/o $S$ & $\Delta$F1 w/o dist. \\
\midrule
HotpotQA & Raw & 0.780 & 0.244 & +0.535 & -0.006 \\
 & SF & 0.806 & 0.191 & +0.615 & -0.010 \\
 & BM25@5 & 0.690 & 0.287 & +0.404 & -0.011 \\
2Wiki & Raw & 0.782 & 0.296 & +0.487 & -0.019 \\
 & SF & 0.822 & 0.255 & +0.566 & +0.004 \\
 & BM25@5 & 0.629 & 0.276 & +0.352 & -0.005 \\
MuSiQue & Raw & 0.581 & 0.093 & +0.487 & -0.044 \\
 & SF & 0.651 & 0.067 & +0.584 & -0.063 \\
 & BM25@5 & 0.377 & 0.136 & +0.241 & -0.012 \\
\bottomrule
\end{tabular}
\end{table}

Matched distractor removal does not reproduce the support-removal drop. Across
raw context, support-first, and BM25@5, deleting annotated support lowers F1 by
0.241--0.615, while deleting matched distractors changes F1 by at most 0.063
and often improves it. The control is therefore consistent with a specific
support-reliance interpretation: the large drops in Table~\ref{tab:supp-reliance-full}
come from removing answer-bearing evidence, not from deleting arbitrary text or
reducing prompt length.

\subsection{Model Size and Cross-Reader Checks}

\begin{figure}[h]
\centering
\includegraphics[width=\columnwidth]{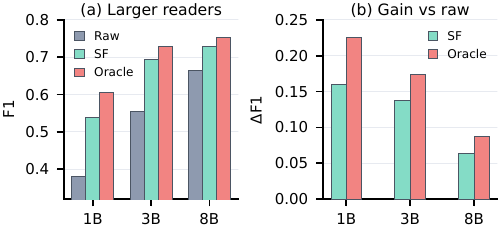}
\caption{HotpotQA larger-reader control at train size 500. Support-first
and oracle-support diagnostics remain above raw context for 1B, 3B, and 8B
readers.}
\label{fig:aaai-model-size}
\end{figure}

Figure~\ref{fig:aaai-model-size} visualizes the HotpotQA larger-reader controls
reported in the main paper. The support-first diagnostic remains above
raw context for 1B, 3B, and 8B readers at train size 500, and the same pattern
holds as the 3B reader is trained with more examples.

The gain is not identical across readers. Larger readers narrow some gaps,
which is expected if they are better at using noisy context. The direction,
however, remains stable: support-first and oracle support stay above raw
context in the measured HotpotQA checks.

Table~\ref{tab:supp-cross-reader-full} lists all measured support-first gains
over raw context. Because model families and train sizes differ across rows,
the table should not be read as a model leaderboard. Its role is directional:
support-first is positive in every measured row.

\begin{table}[h]
\centering
\scriptsize
\setlength{\tabcolsep}{4pt}
\caption{All measured support-first gains over raw context. Values are matched F1 differences; train size differs by row and should not be used as a leaderboard comparison.}
\label{tab:supp-cross-reader-full}
\resizebox{\columnwidth}{!}{%
\begin{tabular}{llrrrr}
\toprule
Dataset & Reader & Train $n$ & Raw & SF & Gain \\
\midrule
HotpotQA & Llama-3.2-3B & 4000 & 0.719 & 0.733 & +0.014 \\
HotpotQA & Qwen-7B & 1000 & 0.696 & 0.748 & +0.052 \\
HotpotQA & Mistral-7B & 1000 & 0.684 & 0.765 & +0.081 \\
HotpotQA & Llama-3.1-8B & 1000 & 0.721 & 0.760 & +0.039 \\
HotpotQA & Gemma-2-9B & 1000 & 0.713 & 0.776 & +0.062 \\
HotpotQA & Qwen-14B & 4000 & 0.780 & 0.806 & +0.026 \\
2Wiki & Llama-3.2-1B & 4000 & 0.475 & 0.523 & +0.048 \\
2Wiki & Llama-3.2-3B & 4000 & 0.657 & 0.732 & +0.075 \\
2Wiki & Qwen-7B & 1000 & 0.618 & 0.650 & +0.032 \\
2Wiki & Mistral-7B & 1000 & 0.585 & 0.700 & +0.115 \\
2Wiki & Llama-3.1-8B & 1000 & 0.582 & 0.697 & +0.115 \\
2Wiki & Gemma-2-9B & 1000 & 0.665 & 0.765 & +0.100 \\
2Wiki & Qwen-14B & 4000 & 0.782 & 0.822 & +0.039 \\
MuSiQue & Llama-3.2-1B & 4000 & 0.299 & 0.459 & +0.159 \\
MuSiQue & Llama-3.2-3B & 4000 & 0.507 & 0.587 & +0.080 \\
MuSiQue & Qwen-7B & 1000 & 0.434 & 0.560 & +0.126 \\
MuSiQue & Mistral-7B & 1000 & 0.404 & 0.506 & +0.102 \\
MuSiQue & Llama-3.1-8B & 1000 & 0.505 & 0.619 & +0.115 \\
MuSiQue & Qwen-14B & 4000 & 0.581 & 0.651 & +0.070 \\
\bottomrule
\end{tabular}
}
\end{table}

The cross-reader table is therefore a robustness check rather than a scaling
law. It shows that the main effect is not tied to one Qwen2.5-Coder-14B checkpoint, while
still leaving open how the effect changes for stronger readers or different
adaptation methods.

\begin{table}[h]
\centering
\scriptsize
\setlength{\tabcolsep}{2.5pt}
\caption{Llama-3.1-8B conventional-reader replication (train size 1000; 1000 matched evaluation examples). Full/missing columns report ranker@5--raw within the corresponding coverage bin. Restore and delete keep the @5 adapter fixed; $*$ marks a 95\% paired interval excluding zero.}
\label{tab:supp-llama-replication}
\resizebox{\columnwidth}{!}{%
\begin{tabular}{lrrrrrrr}
\toprule
Dataset & Raw & @5 & SF & Full gap & Missing gap & Restore & Delete \\
\midrule
HotpotQA & 0.706 & 0.707 & 0.755 & +0.016 & -0.269* & +0.273* & +0.249* \\
2Wiki & 0.623 & 0.618 & 0.710 & +0.068* & -0.113* & +0.208* & +0.328* \\
MuSiQue & 0.463 & 0.360 & 0.574 & +0.064* & -0.179* & +0.263* & +0.248* \\
\bottomrule
\end{tabular}
}
\end{table}

The expanded Llama-3.1-8B check is stronger than a support-first-only
directionality test: it repeats the coverage-conditioned sign reversal and
fixed-adapter mechanism on the same 1000 source examples. Every missing-support
@5 interval is negative, every restore interval is positive, and every delete
drop interval is positive. This substantially reduces dependence of the core
mechanism on the code-tuned main reader, although the Llama adapters use a
smaller train size (1000 rather than 4000).

\FloatBarrier

\section{Case Study and Analysis}

The aggregate diagnostics in the main paper correspond to recognizable
example-level failures. This section expands the support-first win cases used
for Table~\ref{tab:supp-error-taxonomy}. The examples are not an additional
evaluation metric; they explain what the aggregate rows mean when raw context
contains support but still leads the adapted reader to the wrong answer.

\begin{table}[h]
\centering
\scriptsize
\setlength{\tabcolsep}{3pt}
\caption{Support-first win-case taxonomy. Cases are selected where raw context fails and support-first succeeds.}
\label{tab:supp-error-taxonomy}
\begin{tabular}{lrrrr}
\toprule
Dataset & Layout & Entity/hop & Form & Other \\
\midrule
HotpotQA & 4 & 1 & 0 & 2 \\
2Wiki & 10 & 5 & 0 & 2 \\
MuSiQue & 14 & 4 & 0 & 0 \\
\bottomrule
\end{tabular}
\end{table}

The taxonomy is deliberately simple. A \emph{layout} case means that annotated
support is present in the raw candidate pool but appears late, is split by
distractors, or both. An \emph{entity/hop} case means that the raw reader
selects an entity associated with a distractor, a competing candidate, or an
intermediate hop. The \emph{other} bucket covers wrong answers that are not
cleanly explained by these deterministic tags. The table should therefore be
read as an interpretability aid for the quantitative results, not as a final
human error annotation study.

\paragraph{How the interface changes what the reader sees.}
The interfaces are not different question-answering tasks. They are different
renderings of the same retrieved or annotated evidence pool. The following
HotpotQA example illustrates the transformation. Raw context keeps the
candidate documents in their original mixed order. Support-first moves the
complete title-associated support paragraphs to the front and then appends the remaining
distractors. Oracle support removes distractors entirely, while the structured
diagnostic converts the support into a compact relation-like form when the
dataset annotation supports it.

\begin{promptbox}
\textbf{Question:} What government position was held by the woman who
portrayed Corliss Archer in the film \emph{Kiss and Tell}?

\textbf{Raw context sketch:} \emph{Meet Corliss Archer}; \emph{Shirley
Temple}; other distractor documents; \emph{Kiss and Tell}; ...

\textbf{Support-first sketch:} \emph{Kiss and Tell} $\rightarrow$ Shirley
Temple plays Corliss Archer; \emph{Shirley Temple} $\rightarrow$ later served
as Chief of Protocol; then distractors.

\textbf{Oracle-support sketch:} only the two support units above.

\textbf{Gold answer:} Chief of Protocol.
\end{promptbox}

This example is useful because the answer is not obtained by adding external
knowledge or changing the label. The same support chain is instead placed
earlier and grouped locally. The diagnostic question is therefore
whether adaptation teaches the reader to use evidence under a particular
presentation, not whether the benchmark contains the answer somewhere in the
raw candidate pool.

\begin{promptbox}
\textbf{2Wiki comparison rendering:}
For ``Which film came out earlier, Moscow Chill or Khote Sikkey?'', raw context
contains the title order
\emph{Khote Sikkey}; \emph{The Night of Nights}; ...; \emph{XX/XY};
\emph{Moscow Chill}; ...
Support-first presents \emph{Khote Sikkey} and \emph{Moscow Chill} together
before the distractors, turning a scattered evidence search into a local
comparison.

\textbf{MuSiQue chain rendering:}
For ``What place gets the most rain where Sandy High School is?'',
support-first places \emph{Climate of Oregon} next to \emph{Sandy High School}
before the remaining long candidate list. The raw layout instead asks the
reader to connect those units across more surrounding text.
\end{promptbox}

These sketches also explain why the paper keeps support-first separate from
oracle support. Support-first still leaves distractors in the input, so it
tests placement and grouping. Oracle support estimates a cleaner diagnostic
condition in which the reader no longer has to ignore distractors at all. A
large gap between these two rows is evidence that distractor handling remains
part of the problem; a large gap between raw and support-first is evidence
that placement and grouping matter even before distractors are removed.

\paragraph{HotpotQA: support is present, but the wrong supported entity wins.}
One HotpotQA case asks for the lead character in the 1960s sitcom
\emph{Get Smart}, while also requiring evidence about an actress born in 1933.
The raw-context adapter answers \emph{Agent 99}, a plausible entity from the
same show, but the gold answer is \emph{Maxwell Smart}. Support-first answers
the gold string.

\begin{promptbox}
\textbf{Question:} What was the name of the lead character in the 1960s sitcom
``Get Smart'', which also featured an American actress born in 1933?

\textbf{Gold answer:} Maxwell Smart

\textbf{Raw-context answer:} Agent 99

\textbf{Support-first answer:} Maxwell Smart

\textbf{Support titles:} \emph{Get Smart, Again!}; \emph{Barbara Feldon}.
The support titles are separated in the raw order, with distractors between
them.
\end{promptbox}

This is not a missing-evidence failure. The relevant support titles are in the
candidate pool, but the raw interface lets a salient nearby entity dominate the
answer. Support-first helps because it exposes the two support units as the
first chain the reader sees, while still leaving distractors in the input. The
case matches the paper's main distinction: availability alone is too weak, but
oracle support-only is too strong; support-first tests whether the reader can
use the same support when it is easier to encounter.

\paragraph{2WikiMultiHopQA: comparison questions amplify split evidence.}
2WikiMultiHopQA often asks the reader to compare two entities or follow an
entity relation across documents. In one case, the model must decide which of
two films came out earlier. Raw context predicts the competing film,
\emph{Moscow Chill}. Support-first predicts the gold film,
\emph{Khote Sikkey}.

\begin{promptbox}
\textbf{Question:} Which film came out earlier, Moscow Chill or Khote Sikkey?

\textbf{Gold answer:} Khote Sikkey

\textbf{Raw-context answer:} Moscow Chill

\textbf{Support-first answer:} Khote Sikkey

\textbf{Support titles:} \emph{Khote Sikkey}; \emph{Moscow Chill}. One support
unit appears near the beginning of the raw context, while the other appears
much later.
\end{promptbox}

The raw answer is not random: it is one of the two compared entities. This is
why the case is tagged as entity or hop confusion rather than pure layout. The
reader sees both candidates, but their evidence is not presented as a compact
comparison. Moving the support units together changes the interface from a
long mixed context into a local comparison problem. This helps explain why
2WikiMultiHopQA benefits from support-first even when complete support is often
available under larger retrieval windows.

\paragraph{MuSiQue: longer chains make raw order and short windows brittle.}
MuSiQue has more severe support-placement failures because the support units
can be scattered across a longer candidate list. In the following case, raw
context predicts a nearby description, \emph{coastal mountains}, while
support-first predicts the annotated answer, \emph{Coast Range}.

\begin{promptbox}
\textbf{Question:} What place gets the most rain where Sandy High School is?

\textbf{Gold answer:} the Coast Range

\textbf{Raw-context answer:} coastal mountains

\textbf{Support-first answer:} Coast Range

\textbf{Support titles:} \emph{Climate of Oregon}; \emph{Sandy High School}.
The support units are split across a long raw candidate list.
\end{promptbox}

This case illustrates the interaction between coverage and residual reader--interface
analyses. If a top-$k$ window misses one of these support units, the reader does
not have the full chain. If both survive but remain surrounded by distractors,
the reader can still struggle to connect them. MuSiQue therefore has two
failure modes at once: short windows often break the chain, and raw context
often makes the surviving chain hard to use. The main paper therefore treats
retrieval support recall as a prerequisite for interpreting reader-interface
quality.

\paragraph{What the cases add to the aggregate results.}
The cases support three readings of the main experiments. First, support-first
wins are usually not caused by adding new evidence; they come from exposing
evidence that was already present. Second, wrong raw answers are often
semantically close to the support pool, which is why aggregate F1 alone can
hide the mechanism of failure. Third, the same example can be a retrieval
problem under a short top-$k$ window and a reader--interface problem under full
raw context. A practical RAG evaluation should therefore report both support
coverage and reader performance under matched evidence interfaces.

\end{document}